\documentclass[10pt, twocolumn, journal]{IEEEtran}

\usepackage{cite}
\usepackage{graphicx}
\usepackage{epsf}
\usepackage{subcaption}
\usepackage{amsmath}
\usepackage{comment}
\usepackage{amssymb}
\usepackage{array}
\usepackage{setspace}
\usepackage{amsthm,amssymb}
\usepackage[ruled,lined]{algorithm2e}
\usepackage{multirow}
\usepackage{tabularx}
\usepackage{xfrac}
\usepackage{caption}
\usepackage{breqn}
\usepackage{bbm}
\usepackage{enumerate}
\usepackage{algorithmic}
\usepackage{dsfont}
\usepackage{float}
\usepackage{diagbox}
\usepackage{mathtools}
\usepackage{adjustbox}

\graphicspath{{./fig/}}

\title{Attention-Enhanced Deep Learning for Device-Free Through-the-Wall Presence Detection Using Indoor WiFi Systems}
\author{\IEEEauthorblockN{Li-Hsiang Shen$^\dagger$, An-Hung Hsiao, Kuan-I Lu and Kai-Ten Feng}\\
\IEEEauthorblockA{$^\dagger$Department of Communication Engineering, National Central University, Taoyuan, Taiwan\\
Department of Electronics and Electrical Engineering, \\ National Yang Ming Chiao Tung University, Hsinchu, Taiwan\\
gp3xu4vu6@gmail.com, eric28200732.cm07g@nctu.edu.tw, kuani1027@gmail.com, and ktfeng@nycu.edu.tw
}}

\begin{document}

\maketitle

\begin{abstract}
Accurate detection of human presence in indoor environments is important for various applications, such as energy management and security. In this paper, we propose a novel system for human presence detection using the channel state information (CSI) of WiFi signals. Our system named attention-enhanced deep learning for presence detection (ALPD) employs an attention mechanism to automatically select informative subcarriers from the CSI data and a bidirectional long short-term memory (LSTM) network to capture temporal dependencies in CSI. Additionally, we utilize a static feature to improve the accuracy of human presence detection in static states. We evaluate the proposed ALPD system by deploying a pair of WiFi access points (APs) for collecting CSI dataset, which is further compared with several benchmarks. The results demonstrate that our ALPD system outperforms the benchmarks in terms of accuracy, especially in the presence of interference. Moreover, bidirectional transmission data is beneficial to training improving stability and accuracy, as well as reducing the costs of data collection for training. To elaborate a little further, we have also evaluated the potential of ALPD for detecting more challenging human activities in multi-rooms. Overall, our proposed ALPD system shows promising results for human presence detection using WiFi CSI signals.
\end{abstract}


{\let\thefootnote\relax\footnotetext
{Li-Hsiang Shen is with Department of Communication Engineering, National Central University, Taoyuan 32001, Taiwan. (email: gp3xu4vu6@gmail.com)}}

{\let\thefootnote\relax\footnotetext
{An-Hung Hsiao, Kuan-I Lu and Kai-Ten Feng are with the Department of Electronics and Electrical Engineering, National Yang Ming Chiao Tung University (NYCU), Hsinchu, Taiwan. (email: e.c@nycu.edu.tw, kuanil1027@gmail.com and ktfeng@nycu.edu.tw)}}

%%%%%%%%%%%%%%%%%%%%%%%%%%%%%%%%%%%%%%%%%%%%
\section{Introduction}

In recent years, the concept of smart home has gained immense popularity due to its ability to adapt to surroundings and take appropriate actions. Consequently, the detection of human presence has emerged as a pivotal element in a wide range of smart home applications, including but not limited to intruder alarms, elderly home care and emergency warnings, and automatic control of smart home appliances. To make life more convenient and safer, long-term and stable presence detection methods are essential \cite{sur1,sur2,acm}. There emerge numerous schemes achieving presence detection based on existing technologies, which most of them require the deployment of multiple sensors. One of the most popular methods is the camera-based surveillance system \cite{1}. While this system provides highly detailed information about human presence, including body movements and facial features, it also has drawbacks, i.e., privacy cannot be guaranteed, and multiple cameras incur high costs to ensure full-coverage. Infrared sensors \cite{2} are another commonly used commercial presence detection solution, typically operated in a small area such as toilets and stairwells, to control light switches to achieve energy-saving goals. While camera systems raise privacy concerns, infrared sensors offer a more privacy-friendly alternative with relatively low cost. However, there are still problems with infrared sensor-based presence detection systems. For instance, these systems have limitations in detecting stationary individuals and can also experience misdetections in blind spots. Other presence detection methods involve utilizing sound \cite{3} and temperature \cite{4} generated by people. However, these environmental factors fluctuate more significantly than those using cameras and infrared sensors, which lead to severe misdetection.

Due to the limitations mentioned above, the utilization of wireless signals for human presence detection and indoor localization has attracted much research attention. Unlike camera-based systems, wireless signals raise fewer privacy concerns. Thanks to the nature of multipaths from wireless signal reflection, the occurrence of blind spots can be effectively alleviated. For example, radar is considered as a potential technique for indoor sensing. Frequency-modulated continuous-wave radars in millimeter wave frequency bands are adopted in \cite{new0, new1, new2, new3} for creating high-resolution range-Doppler maps. However, they are limited by the measurement distance owing to its higher operating frequencies. Human signatures using radar are detected in \cite{new0}, whilst noninvasive radar-based human activity recognition is performed in \cite{new1, new2}. The radar-based sensing for respiration vital sign of live victim is demonstrated in \cite{new3}. Though with benefits of high-resolution, millimeter wave radar-based sensing accuracy can be severely degraded due to signal blockage. Moreover, Bluetooth beacon \cite{5} relies on mobile phones to analyze the received signal strength indicator (RSSI) sent by pre-installed beacons for sensing. On the other hand, the WiFi system is considered as a promising solution for human presence detection and localization, or even subtle motion-sensing. The channel state information (CSI) from access points (APs) is capable of providing more information than RSSI which can be extracted from the physical layer (PHY) from routers \cite{6}. The CSI information represents the influence of the signal from the transmitter through the channel to the receiver, such as scattering, fading, and energy attenuation with distance. It possesses higher sensitivity level than that of RSSI, which potentially improves the detection accuracy. CSI is widely adopted in various applications, including FiDo \cite{7} and FILA \cite{30} for localization, WiFall \cite{8} and RT-Fall \cite{10} for fall detection, CARM \cite{18}, ABLSTM \cite{25}, and HARNN \cite{24} for human activity recognition, WiGeR \cite{22} and WiFinger \cite{29} for gesture recognition, FarSense \cite{23} and TinySense \cite{28} for respiration sensing, and WRIFI \cite{9} for air-writing recognition. Moreover, by analyzing the amplitude and phase of CSI \cite{11,14,15}, we can obtain its power delay profile, amplitude, angle-of-arrival and time-of-arrival from any pairs of antennas of transmitter and receiver, enabling more information to adopt CSI.

Several studies have been conducted to employ CSI for human presence detection. For example, the work in \cite{pd_svm} has utilized density-based spatial clustering to reduce noise on CSI. Support vector machine (SVM) is applied to classify whether a person is either standing or walking indoors. The authors in \cite{f_lstm} attempt to filter CSI dataset with the aid of moving average filters as well as long short-term memory (LSTM) to detection human presence. Asymptotically, CSI amplitude and phase information are transformed into images by parallel convolutional neural networks (CNNs) \cite{p_cnn}, which extract rich CSI features for detecting an empty room and moving people in a room. The work of \cite{yuming} has proposed a CNN-based autoencoder to reduce the CSI dimensionality as the input into deep neural network for detecting presence in multi-spots.

The majority of prior studies have deployed a pair of WiFi access points (APs) for detection purposes, typically in a small area in a single room \cite{sense_bm1}, which focused on fine-grained signal processing \cite{mycolor}. However, when encountered with presence detection across multiple rooms, signals will be diversified with significant challenges associated with the feature in each room \cite{ourcsichu, ourcsiKI}. In \cite{ourcsichu}, joint dynamic and spatial CSI features are extracted during preprocessing, whereas a recurrent neural network is designed for multi-room presence detection. This requires consideration of multipath attenuation, which can impact detection accuracy. Moreover, even with CSI, there are still some problems to be addressed, i.e., it can be challenging to detect a person standing still in a room due to its subtle change, especially under non-line-of-sight (NLoS). When a stationary person is standing at corner, it may not significantly disturb the wireless channel, causing CSI characteristics to become much similar to those in an empty room. This similarity induces compellingly higher challenges to detect stationary individuals than moving ones. Thus, addressing such problem is crucial to achieve accurate human presence detection. Additionally, if two APs are required to detect a single room, massive APs would be needed for multiple rooms, which is impractical for ordinary households with a compellingly high cost. As the detection coverage area enlarges, it becomes essential to balance deployment costs, implementation complexity as well as detection accuracy.

In this paper, we have conceived a attention-enhanced deep learning for presence detection (ALPD) to address the aforementioned problems. The ALPD system considers bidirectional transmission data for training from a pair of APs, which can compensate the issues caused by deploying two pairs of APs in two rooms. Moreover, the system is designed to extract static and dynamic features that respectively correspond to situations when a person is standing still or walking in the room. For the static feature, attention-based neural network enables the system to emphasize specific subcarriers by assigning appropriate weights to them. The main contributions of this paper are summarized as follows.
\begin{itemize}
	\item We have conceived an ALPD system for device-free presence detection in a two-room scenario. We establish a pair of bidirectional APs capable of either transmitting or receiving signals, creating different CSI data features. We explore various detection cases emcompassing an empty room, the presence of a static individual or a person in motion within one of two rooms, and both rooms occupied by humans. Note that both line-of-sight (LoS) and NLoS scenarios are considered in the static case as well.

	\item ALPD extracts spectral, spatial and temporal features from the collected CSI dataset. We design an attention-based subcarrier weighting mechanism to emphasize the significance of certain informative subcarriers. Joint CNN- and LSTM-based feature extractor is adopted to extract spatial and temporal features in CSI, respectively. Loss function is designed by leveraging reconstruction in autoencoder and clustering losses.

	\item A series of experiments is conducted to observe the CSI using bidirectional APs. We evaluate our proposed ALPD system under various conditions, considering the presence or absence of static/dynamic CSI features as well as the capability of bidirectional APs. We have also evaluated the potential of ALPD for detecting fine-grained human activities in multi-rooms. It reveals that ALPD outperforms the other existing benchmarks in open literature.

\end{itemize}

The rest of this paper is organized as follows. In Section \ref{sys}, we present
the system model and preliminary observations on CSI. The proposed deep learning architecture of ALPD is demonstrated in Section \ref{proposed}. In Section \ref{exp}, experimental trials of ALPD are performed, analyzed, and evaluated. Finally, the conclusions are drawn in Section \ref{con}.

\section{System Model and Experimental Observations} \label{sys}
\subsection{System Model for Through-the-Wall Scenario}
\begin{figure}
\centering
\includegraphics[width=3.5in]{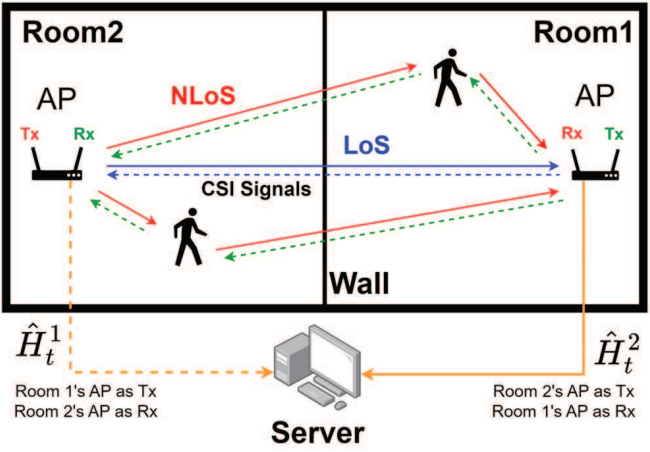}
\caption{\small The scenario of through-the-wall human presence detection. Two rooms are deployed with an AP, where each AP can be operated as either Tx or Rx at one time controlled by the server. The CSI signal includes LoS and NLoS due to multipath effects, which are collected by AP and processed by the server.}
\label{fig:scenario}
\end{figure}

\begin{table}[!t]
\centering
\small
\caption {Four Cases of Through-the-Wall Detection}
\begin{tabular}{|c||c|c|c|c|}
\hline
      & Case 1 & Case 2   & Case 3   & Case 4   \\ \hline
Person in Room 1 &       & $\checkmark$ &         & $\checkmark$ \\ \hline
Person in Room 2 &       &         & $\checkmark$ & $\checkmark$ \\ \hline
\end{tabular}
\label{case}
\end{table}

Our goal is to design a device-free human presence detection system under through-the-wall scenario as shown in Fig. \ref{fig:scenario}. Each of the two adjacent rooms is equipped with a WiFi AP, where both APs are serving as a pair of signal transmitter (Tx) and receiver (Rx). Notice that bidirectional CSI signal transmissions are considered such that each AP will be treated as either Tx or Rx based on the designed scheme. The server will collect the CSI data received by Rx and determine whether there is a human presence in each room at that moment. We classify the experimental through-the-wall scenarios into 4 different cases as shown in Table \ref{case}, where '$\checkmark$' indicates that there is a person either walking or stationary in the room. Note that we consider at most one person existing in a room. This is regarded as the most difficult human presence detection task as only a few multipaths are alternated \cite{my_cronos}. Once there exist more than two people in a room, lots of signal multipaths will be changed, providing simpler human presence detection. CSI is considered the critical information in wireless channel estimation, which can be obtained based on two primary techniques in IEEE 802.11n, i.e., multiple-input multiple-output (MIMO) \cite{20} and orthogonal frequency-division multiplexing (OFDM) \cite{21}. In the MIMO system, the signal from Tx passes through a channel matrix to Rx providing multiple antenna pairs for CSI data. Furthermore, the OFDM technique can offer more channel utilization, where all subcarrier signals are orthogonal to each other within a channel.

In an indoor environment, we can establish the channel model by the MIMO-OFDM system, which can be formulated as 
\begin{align}
    y^p_{t,k,l}=h^p_{t,k,l}x^p_{t,k,l}+n^p_{t,k,l},
\end{align}
where $x^p_{t,k,l}$ and $y^p_{t,k,l}$ are the transmitted and received signal at $k_{th}$ subcarrier of $l_{th}$ antenna pair in $p_{th}$ transmission pair at time $t$, respectively. The parameter $k \in \{1,2,...,K\}$ and $l \in \{1,2,...,L\}$, where $K$ indicates the number of subcarriers in one antenna pair and $L$ is the number of \textit{antenna} pairs in one \textit{transmission} pair. Notice that there can be two concurrent transmission pairs $p\in\{1,2\}$ in our system, which are performed via the bidirectional transmissions between the two WiFi APs. Due to reciprocal direction of APs, equal number of $L$ is obtained in both pairs. The almost concurrent transmissions can be implemented by modifying the firmeware of WiFi AP such that the Tx and Rx pair can alternatively rotate their roles on transmitting and receiving data. Furthermore, $h^p_{t,k,l}$ and $n^p_{t,k,l}$ are the channel response and the additive white Gaussian noise (AWGN), respectively. The CSI at $k_{th}$ subcarrier of $l_{th}$ antenna pair in $p_{th}$ transmission pair at time $t$ can be estimated as
\begin{align}\label{eq:h_t,s,p}
	\hat{h}^p_{t,k,l}=\frac{y^p_{t,k,l}}{x^p_{t,k,l}} = | \hat{h}^p_{t,k,l} | \, e^{j\sin(\angle \hat{h}^p_{t,k,l})},
\end{align}
where $|\hat{h}^p_{t,k,l} |$ represents the CSI amplitude response, and $\angle \hat{h}^p_{t,k,l}$ corresponds to the phase response. In $\eqref{eq:h_t,s,p}$, we can readily estimate the CSI via sending the pilot signals of $x^p_{t,k,l}$. In this paper, we only consider the amplitude part of CSI since the phase part has been studied to be noisy and indescribable \cite{DeepFi}. Furthermore, in order to eliminate the effects caused by power fluctuations from WiFi hardware, we perform normalization on CSI amplitude $|\hat{h}^p_{t,k,l}|$ by
\begin{align}\label{normalization}
	\tilde{h}^p_{t,k,l} = \frac{|\hat{h}^p_{t,k,l}| - \min\limits_{\forall q}| \hat{h}^p_{t,q,l}| }{\max\limits_{\forall q}| \hat{h}^p_{t,q,l}| - \min\limits_{\forall q}| \hat{h}^p_{t,q,l}|}.
\end{align}
With the removal of power variations, this normalization process will allow us to provide more persistent training model during our learning process.

\begin{figure*}
\begin{center}
\subcaptionbox{\label{fig:beforenormalize}}
	{\includegraphics[width=.45\linewidth]{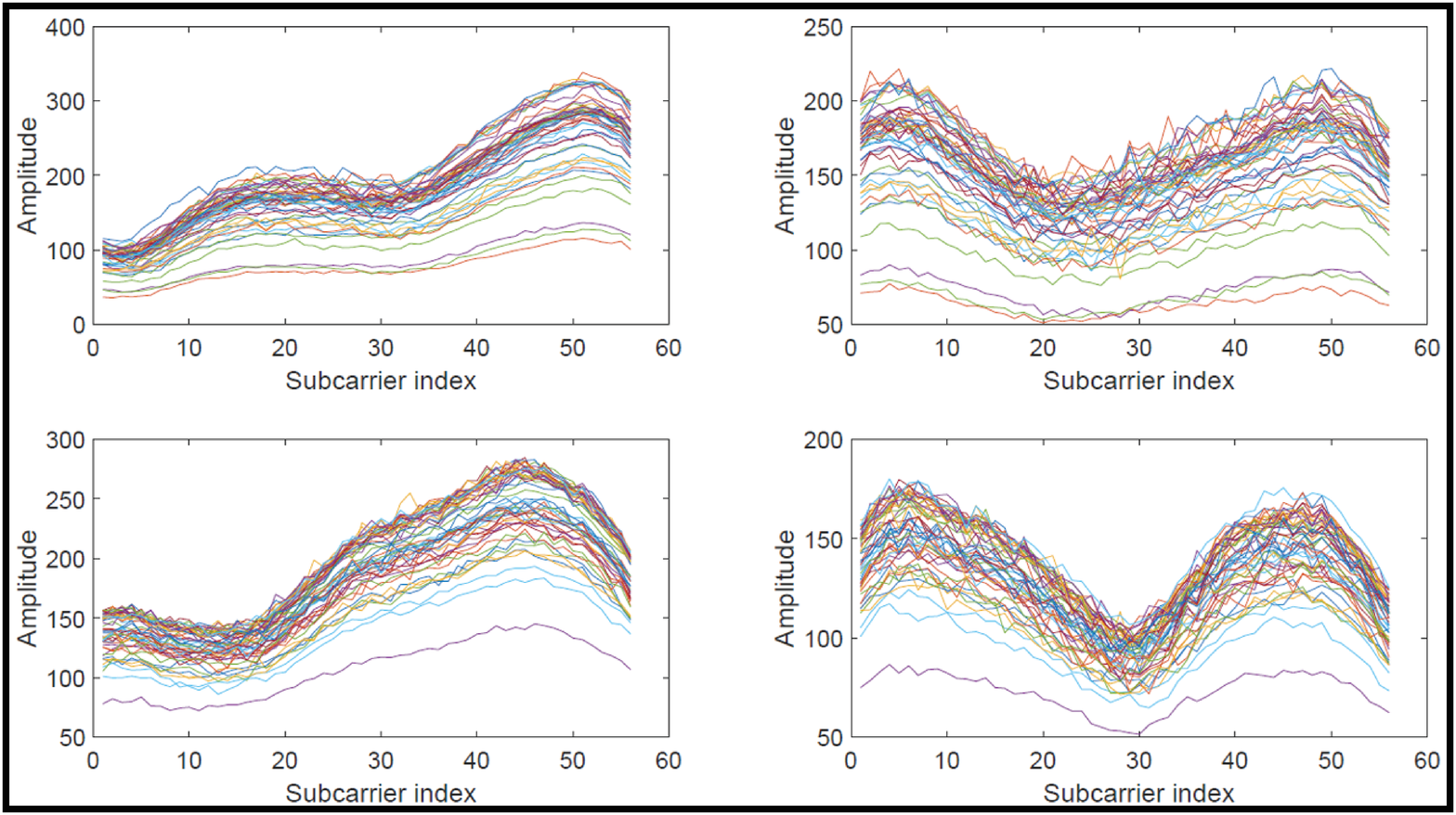}}
	\quad
\subcaptionbox{\label{fig:nopeople}}
	{\includegraphics[width=.45\linewidth]{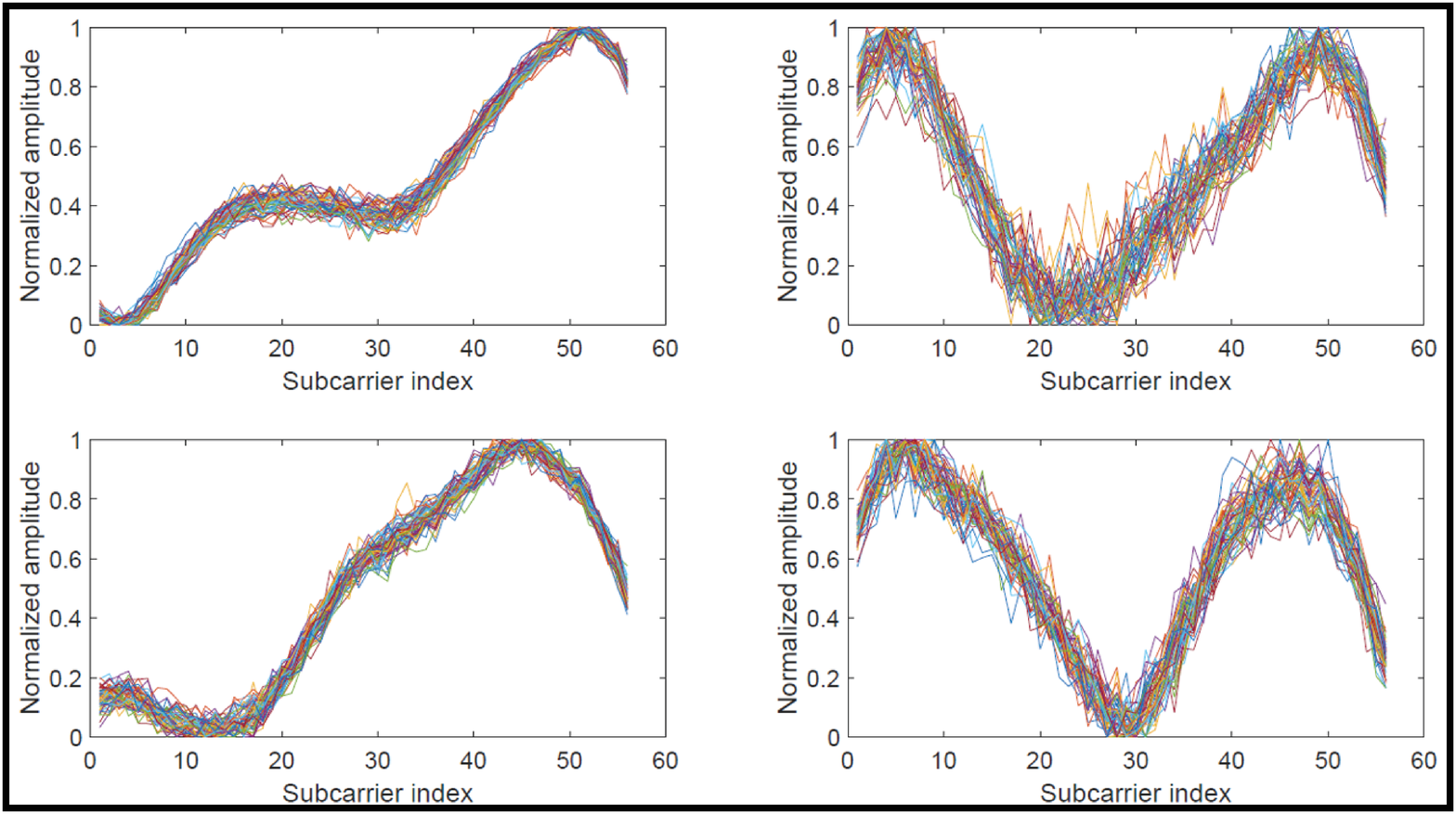}}
\subcaptionbox{\label{fig:d1}}{
	\includegraphics[width=.45\linewidth]{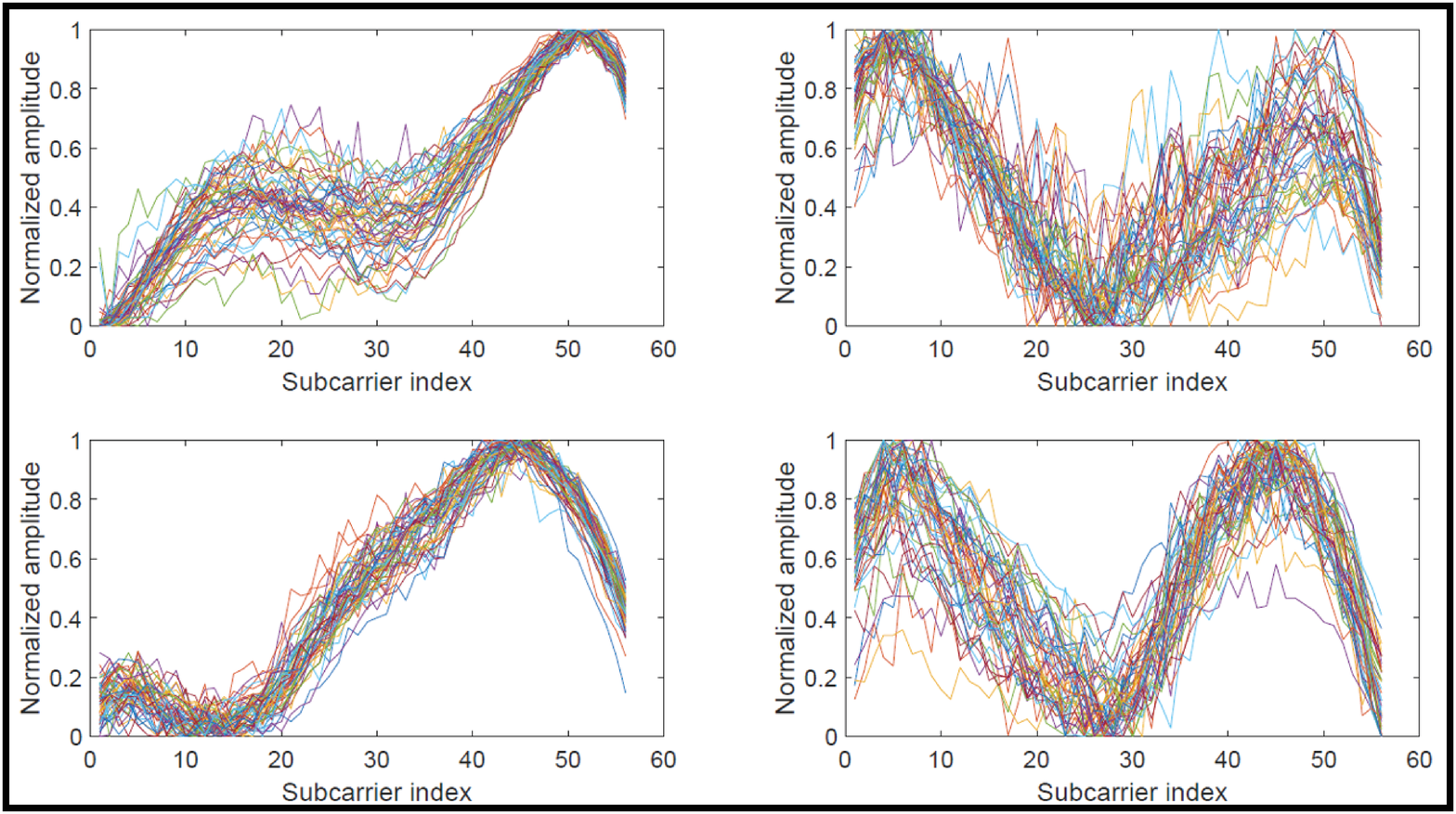}}
	\quad
\subcaptionbox{\label{fig:d2}}{
	\includegraphics[width=.45\linewidth]{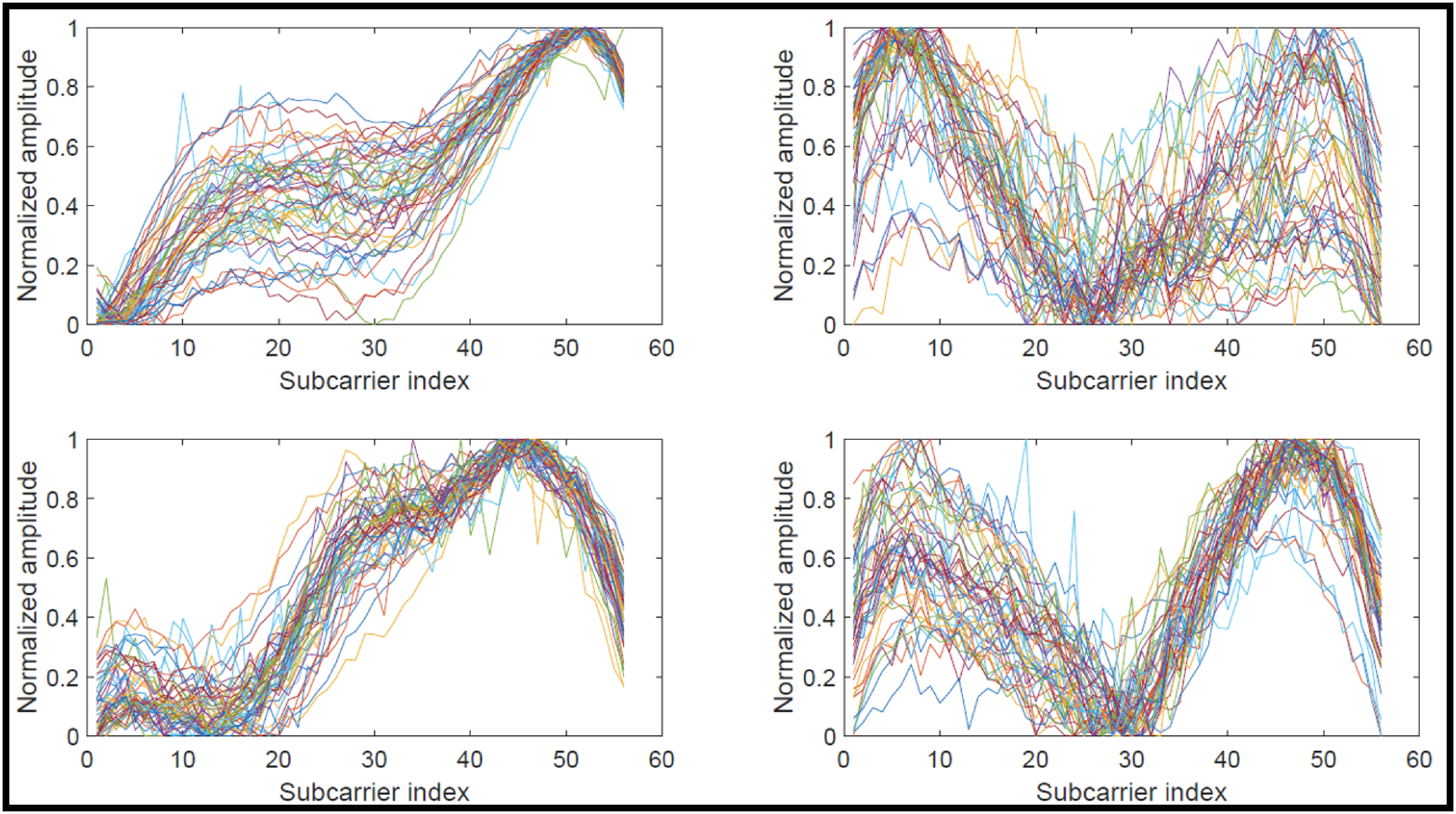}}
\subcaptionbox{\label{fig:sL}}{
	\includegraphics[width=.45\linewidth]{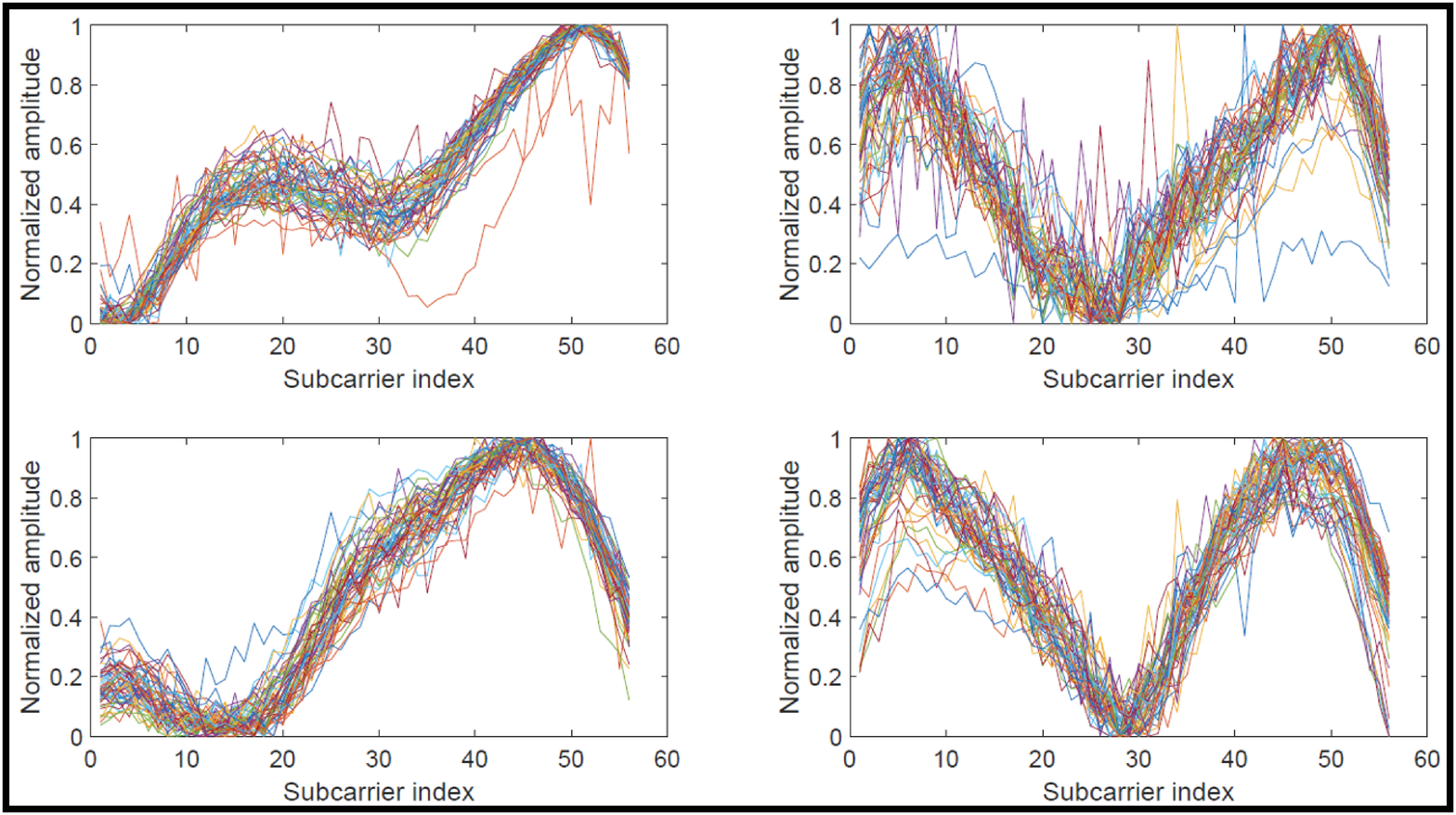}}
	\quad
\subcaptionbox{\label{fig:sN}}{
	\includegraphics[width=.45\linewidth]{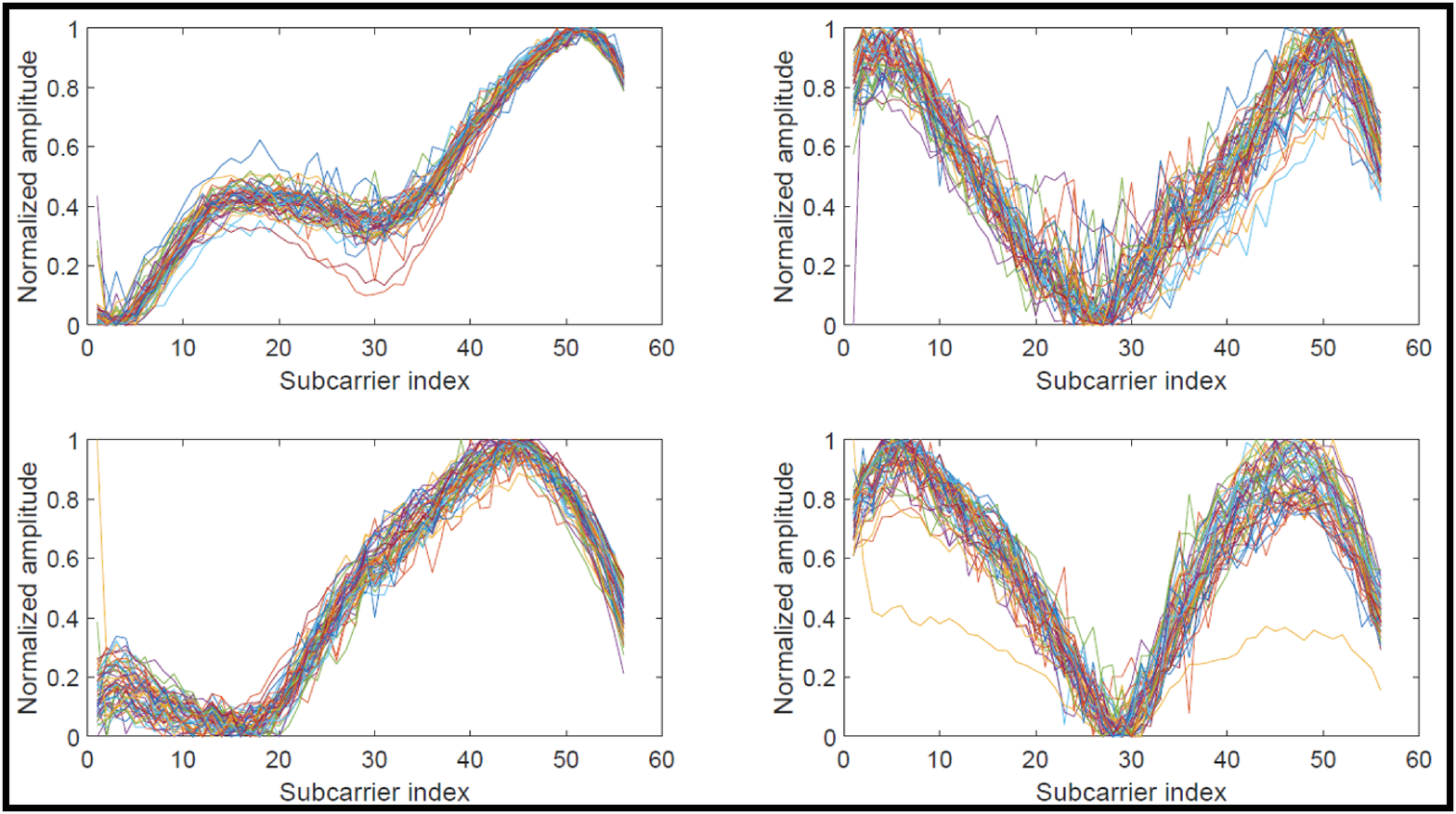}}
\end{center}
\caption{\small Preliminary experimental observations on CSI amplitude versus subcarrier index from four antenna pairs: An empty room case with CSI (a) before and (b) after normalization. The remaining presence cases are collected with normalized CSI, including a person randomly walking in the room deployed with either (c) Tx or (d) Rx, as well as a person standstill at either (e) LoS or (f) NLoS.}
\label{observationFig}
\end{figure*}

\subsection{Preliminary Experimental Observations for Through-the-Wall Scenario}\label{observation}

In order to understand the impact of human presence on wireless signals under through-the-wall scene, we conduct preliminary experiments with the two adjacent rooms as shown in Fig. \ref{fig:scenario}. We deploy two TP-Link TL-WDR4300 routers as Tx and Rx, where each router has two antennas forming four antenna pairs. WiFi APs are set at 2.4 GHz mode, and each antenna pair has 56 subcarriers. We set the transmission rate to 10 Hz and collect 100 CSI packets under different situations, including people walking in either one of the rooms and standing still on either LoS or NLoS scenarios. We intend to observe the effects of human presence causing the changes of wireless multi-path and potential frequency selective fading phenomenon in CSI signals.

\subsubsection{Effects of Normalization Process}
In Fig. \ref{fig:beforenormalize}, we intend to evaluate the effectiveness of normalization process. The result shows the CSI amplitude in an empty room from four antenna pairs under different subcarriers indexed from $k=1$ to $56$, where those curves indicate various data packets from different time intervals. It can be seen that each antenna pair (each subplot) possesses a specific CSI amplitude feature over subcarriers for all different acquired data packets, which can be adopted for identifying various phenomena in the wireless environments. However, as shown in Fig. \ref{fig:beforenormalize}, the automatic power control implemented in WiFi APs results in comparable features across various antenna pairs. In order to eliminate the effects caused by power flucturactions from WiFi hardware, we perform normalization on CSI amplitude $|\hat{h}^p_{t,k,l}|$. After performing the normalization process using $\eqref{normalization}$, we can observe from Fig. \ref{fig:nopeople} that the normalized CSI amplitudes from different timestamps are expected to stay consistent in an empty room. Note that the remaining experimental observations from Figs. \ref{fig:nopeople} to \ref{fig:sN} are performed after the process of normalization.

\subsubsection{Different CSI Patterns between Tx and Rx Sides}

Fig. \ref{fig:nopeople} shows the CSI amplitude over different subcarriers in an empty room where different subcarrier patterns are revealed in four antenna pairs; whereas Figs. \ref{fig:d1} and \ref{fig:d2} illustrate the cases when a person is randomly walking in a room on the Tx and Rx sides, respectively. Comparing with Fig. \ref{fig:nopeople}, it can be easily observed that no matter which room a person is randomly walking around, the CSI data acquired from different time instant fluctuate with people walking around in either one of the rooms. Furthermore, higher fluctuation is shown in Fig. \ref{fig:d2} when people walks in the Rx side compared to that in the Tx side as illustrated in Fig.  \ref{fig:d1}, especially on the top-right antenna pair. The main reason is intuitively due to the wall partition between the two rooms, where the wireless signals received by the Rx from Tx can be blocked, reflected or refracted by the wall. This through-the-wall scenario causes higher signal variations when people walking in the room with Rx compared to that with Tx. In order to validate our speculation, we swap the roles between Tx and Rx such that reversed direction of data transmission is performed and similar experimental results and phenomenons are obtained. Therefore, from this experimental observations, we can obtain different CSI subcarrier patterns among four antenna pairs along with variant fluctuation characteristics between Tx and Rx sides. It will thus be beneficial to preliminarily validate the benefits of conducting bidirectional transmission to compensate the path-loss caused by the through-the-wall scenario. Furthermore, our intent for human presence detection is to identify all these features via our proposed learning scheme which will be explained in the following section.

\subsubsection{Effect when the Person Standing Still} \label{observStill}
As mentioned before, how to provide feasible detection accuracy for a standstill person in an indoor environment has always been a bottleneck in many wireless sensing systems. In order to investigate this situation, Figs. \ref{fig:sL} and \ref{fig:sN} represent the CSI amplitudes when a person is standing still in a room on the Tx side at LoS and NLoS, respectively. Comparing with Figs. \ref{fig:d1} or \ref{fig:d2} with a moving person, it can be easily observed that the fluctuation of CSI for the case with a standstill person is significantly smaller over the subcarrier domain, especially in the NLoS situation as shown in Fig. \ref{fig:sN}. Compared to the case when a person is walking, it is intuitive to speculate that a standstill person has comparably lower impact on the wireless environments since only certain frequencies are disturbed. Therefore, there can be high level of confusion on CSI amplitudes if we intend to distinguish the cases of an empty room and a person standing still at NLoS as revealed in Figs. \ref{fig:nopeople} and \ref{fig:sN}, respectively. This challenging issue should be taken care in our design for human presence detection to provide high detection accuracy.

\section{Proposed Bidirectional Attention-Enhanced Learning Based Presence Detection (ALPD) System}  \label{proposed}

\begin{figure*}
\centering
\includegraphics[width=0.8\textwidth]{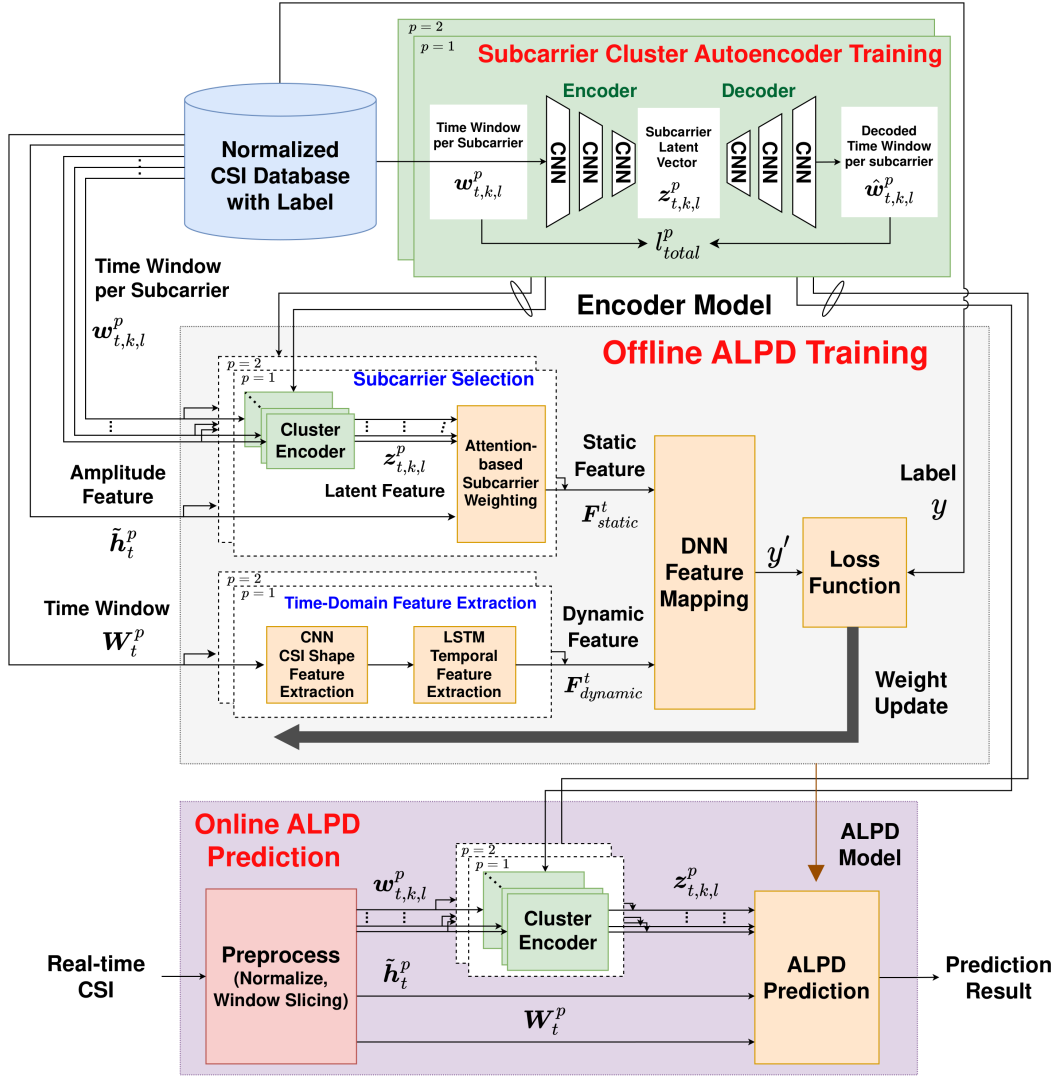}
\caption{\small Schematic diagram of the proposed ALPD system.}
\label{fig:systemmodel}
\end{figure*}

According to the experimental observations of CSI signals in previous section, if we want to achieve high accuracy for indoor presence detection, we must simultaneously handle the situations of a person either walking or standing still in the indoor environments. Moreover, it is beneficial to extract informative CSI amplitudes from the receiver in a bidirectional manner as will be introduced in our proposed ALPD scheme. Fig. \ref{fig:systemmodel} shows the schematic diagrams of our ALPD system, which are divided into offline training and online prediction as follows.

\begin{itemize}
	\item In the offline ALPD training, the amplitude feature from the normalized database will be processed via two paths, including subcarrier selection and time-domain feature extraction in order to respectively extract the static feature (standstill person) and dynamic feature (people walking) in the environments. For the subcarrier selection, we design the attention-based cluster encoder as a processing mechanism to automatically determine the weights for all subcarriers for the extraction of static features. On the other hand, we mainly extract dynamic features from CSI amplitude shape in a time series manner to achieve temporal feature extraction. Finally, as shown Fig. \ref{fig:systemmodel}, both static and dynamic features will be combined to conduct feature mapping by comparing with the labeled data in the database, where the final classified results will be obtained. We can acquire our final ALPD model by continuously updating the corresponding weights and bias.

	\item In the online prediction phase, real-time raw CSI data is processed to obtain the amplitude feature and is organized into the same format as that in the offline stage. We use the same preprocessing method and the previously trained ALPD model to determine the current environmental situation. 

\end{itemize}
In the following subsections, we will first describe our novel design of subcarrier cluster autoencoder (SCAE), and explain in detail the input and output of each block in the proposed ALPD system.

\subsection{Subcarrier Cluster Autoencoder (SCAE)}
In order to extract static features of human presence, our proposed SCAE aims to generate a cluster encoder based on the concept of conventional autoencoder in \cite{13} with enhanced architectural and loss function designs in order to achieve the data clustering purpose. The SCAE can be used to clusterize subcarriers by dimensionality reduction in time-domain. All subcarriers in the same transmission pair will be trained with the same SCAE together. First of all, we consider the temporal characteristics of each subcarrier representing the behavior of specific subcarrier. Therefore, we define the input as the amplitude feature time window of each subcarrier $k$ in antenna pair $l$ in transmission pair $p$ at time $t$ as
\begin{align}
	\boldsymbol{w}^p_{t,k,l} = [\tilde{h}^p_{t-T,k,l},\tilde{h}^p_{t-T+1,k,l},...,\tilde{h}^p_{t,k,l}],
	\label{ww}
\end{align}
where $T$ is the length of considered time window. As shown in Fig. \ref{fig:systemmodel}, the SCAE is composed by both encoder and decoder. After the inputs pass through the encoder, the original $T$-dimension will be reduced to the dimension that we set for the latent vector as
\begin{align}
	\boldsymbol{z}^p_{t,k,l} = \mbox{Encoder}(\boldsymbol{w}^p_{t,k,l}).
\end{align}
After executing the decoder, it will be restored back to the original $T$-dimension as
\begin{align}
	\hat{\boldsymbol{w}}^p_{t,k,l} =  \mbox{Decoder}(\boldsymbol{z}^p_{t,k,l}).
\end{align}

The output of SCAE aims at reconstructing the input so that we can define the total reconstruction loss function $l^p_{recst}$ of transmission pair $p$ as the mean square error (MSE) between input and output as
\begin{align}
	l^p_{recst} = \dfrac{1}{NKL} \sum_{t=1}^{N} \sum_{k=1}^{K} \sum_{l=1}^{L} \mbox{MSE}(\boldsymbol{w}^p_{t,k,l},\hat{\boldsymbol{w}}^p_{t,k,l}),
\label{lr}
\end{align}
where $N$ represents the total number of considered timestamps for data samples of a transmission pair. However, if we only adopt MSE as the total loss function for SCAE, it is difficult to understand what features the latent vector retains due to the reduced dimensionality, i.e., we cannot acquire the basis of clustering. We intend to design the clustering mechanism of SCAE based on the correlation of time window of subcarriers. The correlation of amplitude feature $\boldsymbol{w}^p_{t,k,l}$ between two subcarriers $\{k_i,k_j\}$ in antenna pair $\{l_i,l_j\}$ at time $\{t_i,t_j\}$ can be written as
\begin{align}
\text{cor}(\boldsymbol{w}^p_{t_i,k_i,l_i},\boldsymbol{w}^p_{t_j,k_j,l_j}) = \frac{\text{cov}(\boldsymbol{w}^p_{t_i,k_i,l_i},\boldsymbol{w}^p_{t_j,k_j,l_j})}{\sigma_{\boldsymbol{w}^p_{t_i,k_i,l_i}}\sigma_{\boldsymbol{w}^p_{t_j,k_j,l_j}}},
\label{cor}
\end{align}
where $\text{cov}(\cdot,\cdot)$ represents the covariance function. $\sigma_{\boldsymbol{w}^p_{t_i,k_i,l_i}}$ and $\sigma_{\boldsymbol{w}^p_{t_j,k_j,l_j}}$ are the standard deviations of $\boldsymbol{w}^p_{t_i,k_i,l_i}$ and $\boldsymbol{w}^p_{t_j,k_j,l_j}$, respectively. If two subcarriers are highly correlated in time-domain, it indicates that those two subcarriers possess similar behavior. We intend to assign the corresponding latent vectors $\boldsymbol{z}^p_{t_i,k_i,l_i}$ and $\boldsymbol{z}^p_{t_j,k_j,l_j}$ of those two subcarriers to be closer in the latent space as shown in Fig. \ref{fig:systemmodel}. Therefore, we propose the total clustering loss $l^p_{clst}$, which is defined as
\begin{align}
l^p_{clst} = 
\sum_{\forall \xi_i, \xi_j\ne i}\frac{d(\boldsymbol{z}^p_{t_i,k_i,l_i},\boldsymbol{z}^p_{t_j,k_j,l_j})}{\text{cor}^{+}(\boldsymbol{w}^p_{t_i,k_i,l_i},\boldsymbol{w}^p_{t_j,k_j,l_j}) },
\label{lc}
\end{align}
where $\xi_i = (t_i,k_i,l_i)$ and $\xi_j = (t_j,k_j,l_j)$ are tuples of indexes. Function of $d(\cdot,\cdot)$ is a distance metric between two vectors, which is implemented as the Euclidean distance. The denominator $\text{cor}^{+}(\boldsymbol{w}^p_{t_i,k_i,l_i},\boldsymbol{w}^p_{t_j,k_j,l_j})$ in $\eqref{lc}$ is equal to $\text{cor}(\boldsymbol{w}^p_{t_i,k_i,l_i},\boldsymbol{w}^p_{t_j,k_j,l_j})$ in $\eqref{cor}$ if $\text{cor}(\boldsymbol{w}^p_{t_i,k_i,l_i},\boldsymbol{w}^p_{t_j,k_j,l_j})>0$; otherwise $\text{cor}^{+}(\boldsymbol{w}^p_{t_i,k_i,l_i},\boldsymbol{w}^p_{t_j,k_j,l_j})=0$. Notice that the intent of clustering loss in $\eqref{lc}$ is to consider only positive correlation between the two subcarriers $\boldsymbol{w}^p_{t_i,k_i,l_i}$ and $\boldsymbol{w}^p_{t_j,k_j,l_j}$ in order to enforce their corresponding latent vectors $\boldsymbol{z}^p_{t_i,k_i,l_i}$ and $\boldsymbol{z}^p_{t_j,k_j,l_j}$ to be closer with each others. With higher positive correlation among original subcarriers, the distances among their corresponding latent vectors become smaller which will result in comparably smaller clustering loss for achieving high detection accuracy.

As a result, the total loss function for proposed SCAE is defined as the combination of reconstruction loss in (\ref{lr}) and clustering loss in (\ref{lc}) as 
\begin{align}
l^p_{total} = l^p_{recst} + \lambda l^p_{clst},
\label{lt}
\end{align}
where $\lambda$ is the weighting parameter to balance between $l^p_{recst}$ and $l^p_{clst}$. After the SCAE is successfully trained, we will take out the encoder part of SCAE and name it as {\it cluster encoder}, which means that it will perform the action to feasibly cluster subcarriers based on our designed loss function in $\eqref{lt}$. As illustrated as green boxes of Fig. \ref{fig:systemmodel} in both offline and online APLD phases, the cluster encoder is adopted for subcarrier selection, which will be described in more details as follows.

\subsection{Offline ALPD Training}

In the offline ALPD phase as shown in Fig. \ref{fig:systemmodel}, the training model includes subcarrier selection, time-domain feature extraction, and feature mapping in order to tackle both stationary and moving situations for human presence detection. Note that there will be two transmission pairs because our system is designed to adopt bidirectional transmission data. Data from two transmission pairs independently and parallelly undergo the same processing in both subcarrier selection and time-domain feature extraction functions. These two pairs will be merged together in the feature mapping block to obtain the final estimate.

\subsubsection{Subcarrier Selection}
\begin{figure}
\centering
\includegraphics[width=3.5in]{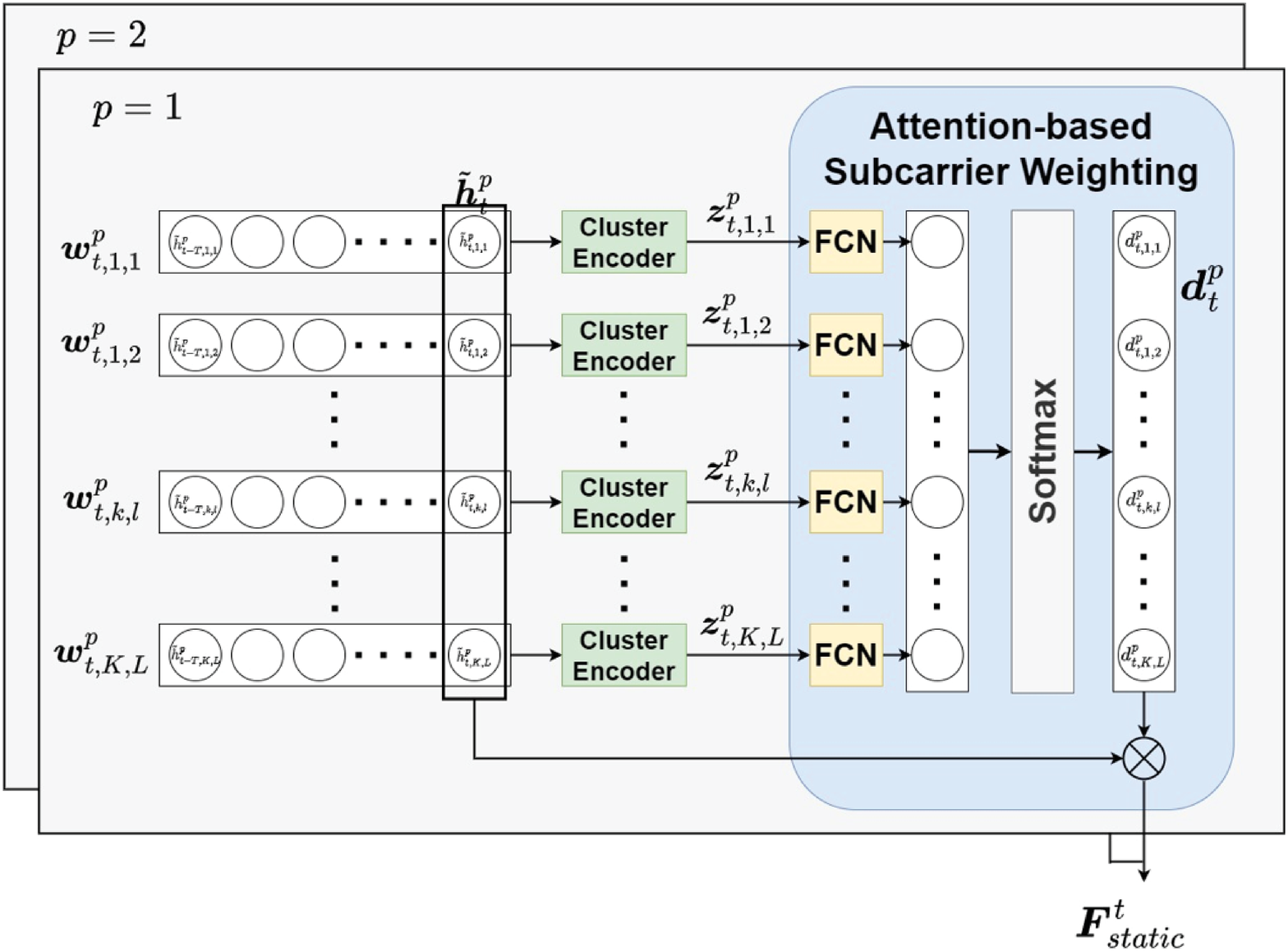}
\caption{\small Architecture of attention-based subcarrier selection.}
\label{fig:subcarrierselection}
\end{figure}

The input/output relationship for subcarrier selection can be seen in Fig. \ref{fig:systemmodel}; whereas the detailed architectural design of subcarrier selection is illustrated in Fig. \ref{fig:subcarrierselection}. There are two types of inputs for subcarrier selection, including $\tilde{\boldsymbol{h}}^p_t$ and $\boldsymbol{w}^p_{t,k,l}$, where 
\begin{align}
\tilde{\boldsymbol{h}}^p_t = [\tilde{h}^p_{t,1,1},\tilde{h}^p_{t,1,2},...,\tilde{h}^p_{t,k,l},...,\tilde{h}^p_{t,K,L}]
\label{h_tilda}
\end{align}
denotes the normalized amplitude feature vector at current time $t$ with its vector elements from (\ref{normalization}) and $\boldsymbol{w}^p_{t,k,l}$ is the amplitude feature time window of $T$ of subcarrier $k$ in antenna pair $l$ as defined in (\ref{ww}). 

As described in Subsection \ref{observStill}, it is challenging to perform human presence detection between the cases of an empty room and a standstill person at NLoS due to their small difference on CSI amplitudes. In order to alleviate the ambiguity problem for presence detection, we intend to conduct micro design to assign different weights for each subcarrier of specific antenna pair. The main design idea of our subcarrier selection is to provide the amplitude feature vector $\tilde{\boldsymbol{h}}^p_t$ at current time $t$ with feasible weight based on $\boldsymbol{w}^p_{t,k,l}$ with the time window of $T$. Instead of using equal weight for all subcarriers, the main reason to apply $\boldsymbol{w}^p_{t,k,l}$ on each subcarrier is that we intend to amplify the amplitude features of specific subcarriers/antenna pairs such that we can enlarge the difference of signal characteristics among different detection scenarios. With the specially assigned weighting values for all subcarriers, we can pay more attention to certain subcarriers to assist improving the detection accuracy.

As shown in Fig. \ref{fig:subcarrierselection}, the input $\boldsymbol{w}^p_{t,k,l}$ is the time characteristics which can be regarded as the behavior of each specific subcarrier. At time $t$, all $\boldsymbol{w}^p_{t,k,l}$ belonging to the same transmission pair $p$ will be individually passed through their specific cluster encoder generated from SCAE after evaluated by the total loss function in $\eqref{lt}$ as described in previous subsection. The output $\boldsymbol{z}^p_{w,k,l}$ can therefore be generated representing the corresponding lower-dimensional clustering result for each subcarrier.

After clustering process, all $\boldsymbol{z}^p_{w,k,l}$ and $\tilde{\boldsymbol{h}}^p_t$ will enter the attention-based subcarrier weighting block together to generate the static feature as illustrated in Fig. \ref{fig:subcarrierselection}. The clustering characteristics for a subcarrier $\boldsymbol{z}^p_{w,k,l}$ first pass through a fully-connected neural network (FCN) to be reduced to a single dimension. These dimensions are then combined into a vector called the raw weighting vector. The raw weighting vector passes through the softmax function, which ensures that the sum of the vector equals $1$. The result is the weight vector, which is the final output as
\begin{align} 
	\boldsymbol{d}^p_t = [d^p_{t,1,1},d^p_{t,1,2},...,d^p_{t,1,L},d^p_{t,2,1},...,d^p_{t,K,L}],
	\label{dd}
\end{align}
where $d^p_{t,k,l}$ represents the final weight corresponding to $\tilde{h}^p_{t,k,l}$. As a result, the subcarrier weighting can be determined by performing element-wise product of $\boldsymbol{d}^p_t$ and $\tilde{\boldsymbol{h}}^p_t$. The static feature at time $t$ is the concatenation of all subcarrier selection results of all transmission pairs, which can be written as
\begin{align}
	\boldsymbol{F}^t_{static} = [\tilde{h}^1_{t,1,1}d^1_{t,1,1},...,\tilde{h}^p_{t,k,l}d^p_{t,k,l},...,\tilde{h}^P_{t,K,L}d^P_{t,K,L}],
\end{align}
where $P$ is the total number of transmission pairs.

\subsubsection{Time-Domain Feature Extraction}
\begin{figure}
\centering
\includegraphics[width=3.5in]{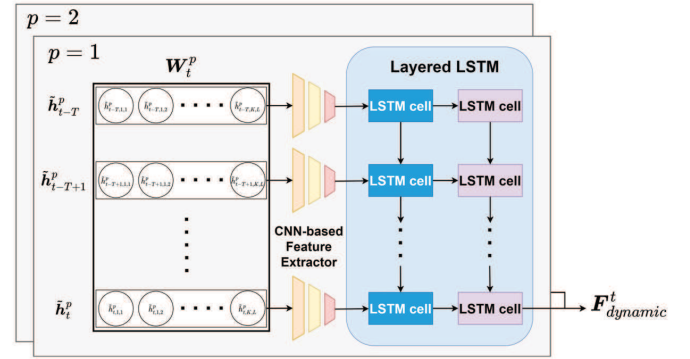}
\caption{\small Architecture of time-domain feature extraction. Note that three-layer CNN and two-layer LSTM are considered for extracting spatial and temporal features from CSI signals.}
\label{fig:TDFE}
\end{figure}
%In this section, we will first introduce Long Short-Term Memory (LSTM) network, which is the most important network for the time-domain feature extraction block. Then cooperate with the Convolutional neural network (CNN) to extract the dynamic feature.

The purpose of time-domain feature extraction is to obtain dynamic characteristics of a person moving in a space. The recurrent neural network (RNN) \cite{27} is a well-adopted model to deal with time series forecasting, which can be a feasible solution to predict human presence based on CSI amplitude inputs. However, RNN is known to suffer from vanishing and exploding gradient problems along with difficulty to update the weight of previous layer's network. Especially, the challenging task of our considered problem is to capture long-term temporal correlation from the CSI signal input in order to extract human moving behaviors. Therefore, the LSTM network proposed in \cite{12} as an advanced variant of RNN can be adopted as a feasible solution to process and predict important events with longer intervals and delays.

Fig. \ref{fig:TDFE} shows the architectural diagram of time-domain feature extraction, which is a zoom-in diagram of that in Fig. \ref{fig:systemmodel}. The inputs to this function are defined as the combination of the amplitude feature vector $\tilde{\boldsymbol{h}}^p_{t}$ within time window $T$ in (\ref{h_tilda}), which can be written as
\begin{align}
\boldsymbol{W}_t^p=[\tilde{\boldsymbol{h}}^p_{t-T},\tilde{\boldsymbol{h}}^p_{t-T+1},...,\tilde{\boldsymbol{h}}^p_{t}].
\end{align}
First of all, the amplitude feature vector $\tilde{\boldsymbol{h}}^p_{t}$ will be passed through the share weights three-layer CNN to extract the {\it shape features} of CSI amplitude before executing LSTM. The reason to implement the three-layer CNN is to potentially distinguish a variety of CSI shape features such as to promote the effectiveness of LSTM on extracting {\it time series features} of dynamic characteristics. After executing the CNN, a two-layer LSTM is performed as shown in Fig. \ref{fig:TDFE} and the corresponding outputs of all transmission pairs are concatenated to form the dynamic feature $\boldsymbol{F}_{dynamic}^t$.

\subsubsection{Feature Mapping}
As shown in Fig. \ref{fig:systemmodel}, we can obtain both static feature $\boldsymbol{F}_{static}^t$ and dynamic feature $\boldsymbol{F}_{dynamic}^t$ based on the above two subsections, which can respectively provide effective feature extraction to detect if a person is standstill or dynamically moving in one of the rooms, especially under through-the-wall scenario. In this feature mapping stage, these two features will be concatenated together and are sent to the series of fully connected neural network, i.e., DNN as illustrated in Fig. \ref{fig:systemmodel}. The model automatically learns the characteristics of these two features and maps them to the corresponding situations, i.e., to conduct classification for presence detection including four label classifiers as empty room, left-room presence, right-room presence, and both-room human presence. By comparing the labels $y$ in the database and predicted output $y'$, we can update the weight and bias using classification loss of cross-entropy.

\subsection{Online ALPD Prediction}

After the offline ALPD model has been established, the online prediction process can be performed for human presence detection. As shown in Fig. \ref{fig:systemmodel}, the real-time measured CSI signals are acquired and the corresponding amplitude features are pre-processed via normalization and window slicing to obtain various types of inputs for proposed ALPD system. After receiving the CSI, we first extract the amplitude feature and then form the time window as the input of the ALPD system. The amplitude feature time window of each subcarrier $\boldsymbol{w}^p_{t,k,l}$ will be passed to the designed cluster encoder to generate the corresponding cluster latent vector $\boldsymbol{z}^p_{t,k,l}$. Three different types of inputs, including the cluster latent vector $\boldsymbol{z}^p_{t,k,l}$, the normalized amplitude feature vector $\tilde{\boldsymbol{h}}^p_t$, and the amplitude feature $\boldsymbol{W}_t^p$ for time window $T$, are provided to perform the ALPD prediction process based on the ALPD model constructed in the feature mapping stage as stated in previous subsection. As a result, the real-time APLD system can be implemented to detect human presence in the through-the-wall environments. To elaborate a little further, this two-room scenario and algorithm design in this work can be readily extended to that for multi-rooms, e.g., two pairs of APs in a three-room case. Each pair can independently operate its algorithm for human presence detection. Therefore, a voting mechanism can be designed for presence detection in the scenario with more than three rooms, which is left as future design. 

\begin{table}[]
\small
\centering
\caption{Parameter Setting}
\begin{tabular}{ll}
\hline
Parameter  	& Value  
\\ \hline \hline
Carrier frequency   	& 2.447 GHz       \\
Channel bandwidth     	& 20 MHz          \\
Number of subcarrier 	& 56              \\
Number of antenna pair 	& 4               \\
Number of WiFi AP    	& 2 \\
Number of transmission pair   	& 2 \\
Data collection rate   	& 10 packets/sec  \\
Training data amount per case    & 10000  \\
Testing data amount per case    & 5000  \\
Length of latent vector & 5 \\
Weight of $\lambda$		& 0.01	\\
\hline
\end{tabular}
\label{systemPar}
\end{table}

\section{Performance Evaluation} \label{exp}

\begin{figure}
\centering
\includegraphics[width=3.5in]{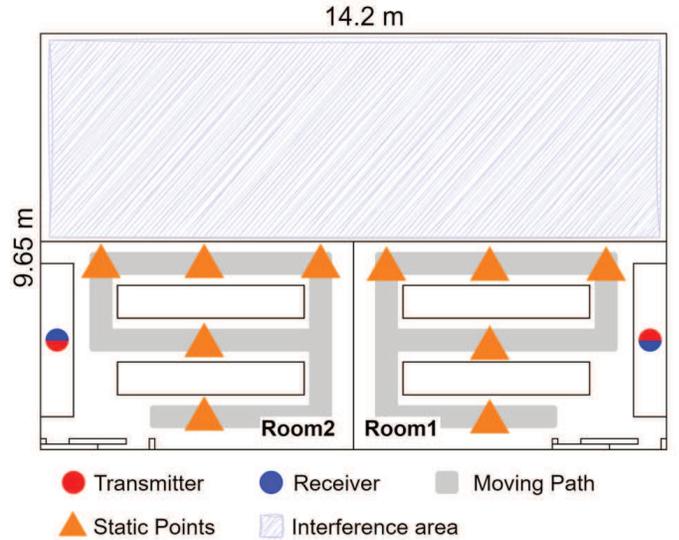}
\caption{\small Experimental scenario and setting.}
\label{fig:testbed}
\end{figure}

\subsection{Experimental Setup}
Field experiments are performed to evaluate the effectiveness of proposed ALPD system\textsuperscript{\ref{note1}}\footnotetext[1]{In this paper, we choose a relatively clean WiFi channel by neglecting the complex interference scenario. When encountering massive WiFi routers, the selected subcarrier feature might be polluted with compelling interference owing to limited channel resources. However, these conditions highly depend on the deployment of WiFi routers and coverage to be sensed. Some other mechanisms should be designed, including interference mitigation, scheduling, or multi-WiFi collaborative sensing, which can be regarded as future works.\label{note1}}. Fig. \ref{fig:testbed} shows our experimental scenario, where the meeting room is separated by partitions into three spaces, including Room 1, Room 2, and the interference area. Our experiments are conducted in both Room 1 and Room 2 to evaluate the through-the-wall scenarios. The interference area, which has several people walking around at random locations, is used to evaluate the detection accuracy by examming the impact of interference outside the testing rooms. In order to implement the bidirectional transmission scenarios in our proposed ALPD system, we adopt two pairs of Tx and Rx (i.e., $p=2$) in the experimental scene. In other words, the CSI signals acquired from both the transmission paths Tx$_1$-Rx$_2$ and Tx$_2$-Rx$_1$ will be collected by the proposed ALPD system at the backend computing platform. Note that Tx$_1$ indicates Tx in Room 1, whilst Rx$_2$ means Rx in Room 2\textsuperscript{\ref{note2}}\footnotetext[2]{We deploy APs nearby the wall of the respective rooms in order to capture the maximum detection coverage. To elaborate further, deployment is a critical issue in wireless sensing applications. Different deployments of APs potentially affect the sensing performance. For instance, placing the APs too close with each other will induce the dominant LoS signals, whereas results with faraway APs will be dominated by the pathloss attenuation. Moreover, deploying the AP in a spot with plenty of undesirable objects, the performance will be influenced by the blockage effects or weak multi-path signals. It would be time-consuming for conducting fingerprinting for different deployment locations, just list a few examples. Optimal AP deployment should further consider lots of factors, which is regarded as an open issue in sensing applications.\label{note2}}. Four TL-WDR4300 wireless routers operated in IEEE $802.11$n protocol are deployed in our experiments serving as two transmitters and two receivers. There are two antennas on each router, which results in four antenna pairs in one transmission path, i.e., $l=4$. All routers are operated in central carrier frequency $2.447$ GHz with $20$ MHz bandwidth with corresponding $56$ subcarriers in one antenna pair. For example, the CSI at $k=56_{th}$ subcarrier of $l=4_{th}$ antenna pair in $p=1_{th}$ transmission pair, i.e., Tx$_1$-Rx$_2$, at time $t$ can be represented as $\hat{h}^p_{t,k,l}=\hat{h}^1_{t,56,4}$ in (\ref{eq:h_t,s,p}).

The data collection rate is set as $10$ Hz. We collect a total of $10000$ data for each case at the training stage and $5000$ data for each case in testing phase, where the interval between the two collection phases is one hour to exam the influence from different time intervals. For each case, the tester is freely walking around in the moving path and standing still on any static points at any time, which is defined as {\it normal state} in our experimental testings. In addition, we also collect two special types of testing data defined as {\it static state} and {\it interfering state}, each with the data amount of $5000$. The purpose of examining the static state is to evaluate a person standing still on each static point as shown in Fig. \ref{fig:testbed}, i.e., no movements of people occur in this state. While, the interfering state has the same moving/static scenarios as normal testing data but with people walking around in the interference area. Notice that the static state is specially evaluated since it is considered the challenging scenario to be correctly detected in indoor human presence detection. In this experiment, the length of the latent vector is set to be $5$, and parameter $\lambda$ is $0.01$.

\subsection{Effect of Different Parameters}

First of all, we need to determine the feasible length of time window $T$ defined in \eqref{ww} for our proposed ALPD system. The time window size $T$ directly influences the size of amplitude feature time window $\boldsymbol{w}^p_{t,k,l}$ in \eqref{ww}. Table \ref{fig:timestep} shows the detection accuracy $\gamma$ corresponding to different lengths of time window $T$ by adopting testing data at normal state. We can observe that the detection accuracy is increased from $88.5\%$ to $96\%$ with window size enlarged from $T=1$ to $70$. That is, with longer time window size $T$, more data feature will be collected in a considered time window, which can result in higher detection accuracy. However, higher computational complexity will be incurred in our proposed ALPD system along with elongated detection delay. On the contrary, with shorter window size, the ALPT scheme may not be able to provide feasible judgments, resulting in lowered detection accuracy. Therefore, it is necessary to strike a compelling balance on the determination of time window size $T$. Moreover, the detection accuracy saturate at around $T=50$ to achieving $\gamma=95.8\%$. Accordingly, considering both detection accuracy and computational cost, we set the length of the time window to $T=50$ in the following experiments.

In Table \ref{fig:lam}, we evaluate the performance of different weights of $\lambda$. We can observe that it achieves the highest accuracy of $\gamma=95.9 \%$ with the parameter $\lambda=0.001$. This is due to the reason that smaller $\lambda<0.001$ will more focus on the reconstruction of amplitude features, whereas larger $\lambda>0.001$ places a greater emphasis on the feature correlation of clustering. Accordingly, we set $\lambda=0.001$ in the following experiments. Moreover, in Table \ref{fig:len} we evaluate different lengths of latent vector $\boldsymbol{z}^p_{t,k,l}$.  We can infer that insufficient length will provide incomplete hidden feature information for human presence detection, leading to low accuracy, i.e., only $63.81\%$ with latent length of $3$. Though providing more informative features, longer latent lengths will induce comparably higher computational complexity in deep learning training. In the following experiments, we select latent vector length of $5$ since it strikes a compelling balance between complexity and asymptotic accuracy compared to the lengths of $\{10,20\}$.

\begin{table}[!t]
\centering
\begin{footnotesize}
\caption {Accuracy of Different Time Windows}
\begin{tabular}{|l||l|l|l|l|l|l|l|l|}
\hline
\begin{tabular}[c]{@{}l@{}}$T$\end{tabular} & 1    & 10   & 20   & 30   & 40   & 50   & 60   & 70   \\ \hline
$\gamma$ (\%)                                                               & 88.5 & 90.2 & 91.6 & 92.5 & 94.5 & \textbf{95.8} & \textbf{95.8} & \textbf{96.0} \\ \hline
\end{tabular}
\label{fig:timestep}
\end{footnotesize}
\end{table}

\begin{table}[!t]
\centering
\begin{footnotesize}
\caption {Accuracy of Different $\lambda$}
\begin{tabular}{|l||l|l|l|l|l|l|l|}
\hline
\begin{tabular}[c]{@{}l@{}} $\lambda$ \end{tabular} & 0.0001 &0.001 &0.001 &0.1 &1 &10 &100 \\ \hline
$\gamma$ (\%)                                                               & 80.4& 82.51 & \textbf{95.9} & 80.01 & 73.73 & 72.9 & 58.04 \\ \hline
\end{tabular}
\label{fig:lam}
\end{footnotesize}
\end{table}

\begin{table}[!t]
\centering
\begin{footnotesize}
\caption {Accuracy of Different Lengths of Latent Vector}
\begin{tabular}{|l||l|l|l|l|l|}
\hline
\begin{tabular}[c]{@{}l@{}}Length\end{tabular} & 1 & 3 & 5 & 10 & 20 \\ \hline
$\gamma$ (\%)                                                               & 62.72 & 63.81 & \textbf{95.9} & \textbf{95.92} & \textbf{96.17} \\ \hline
\end{tabular}
\label{fig:len}
\end{footnotesize}
\end{table}

\subsection{Evaluation of Subcarrier Selection}

As the main contribution of this work, the designed subcarrier selection SCAE scheme is responsible for dealing with the situation of the person standing still in the room for through-the-wall scenario. Since the performance of proposed SCAE scheme will significantly affect the presence detection accuracy, we will analyze it step-by-step in the following subsections.

\begin{figure}
\centering
\subcaptionbox{\label{fig:watt2}}{
\includegraphics[width=0.48\linewidth]{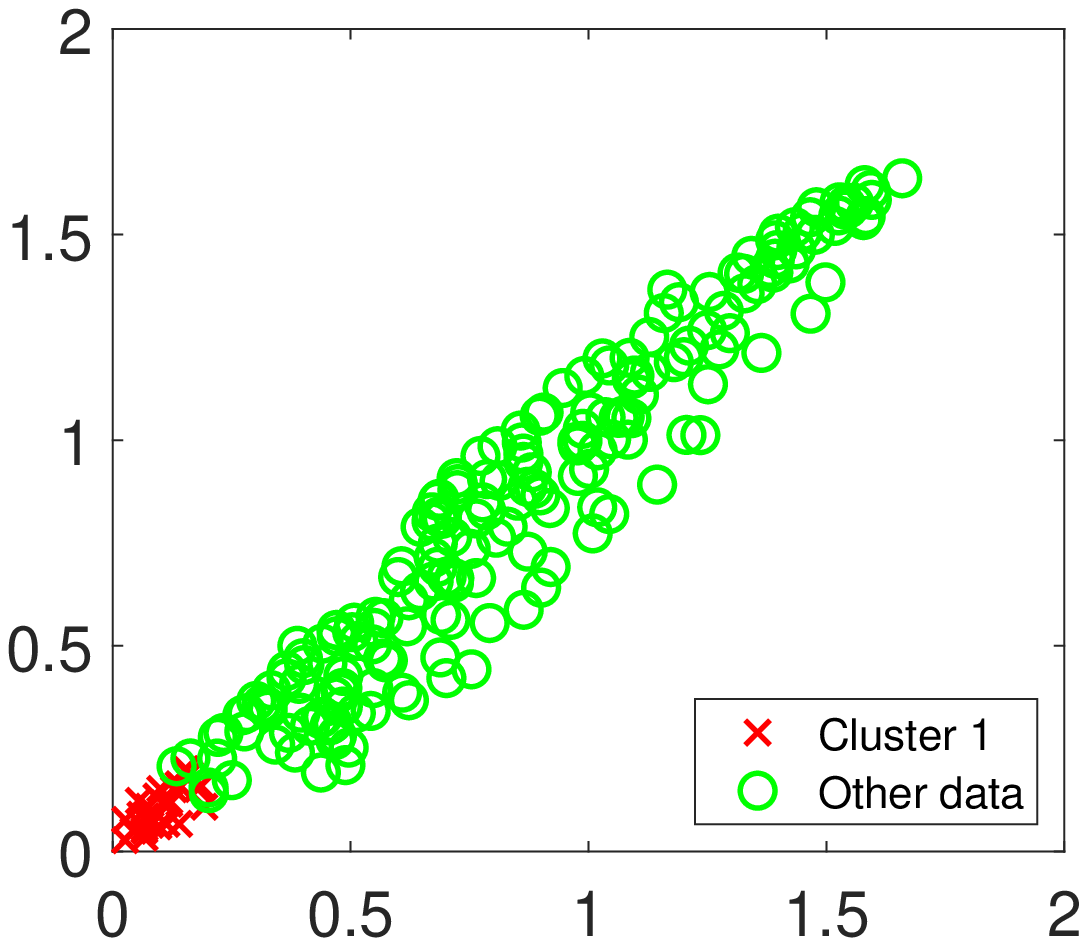}}
\subcaptionbox{\label{fig:woatt2}}{
\includegraphics[width=0.48\linewidth]{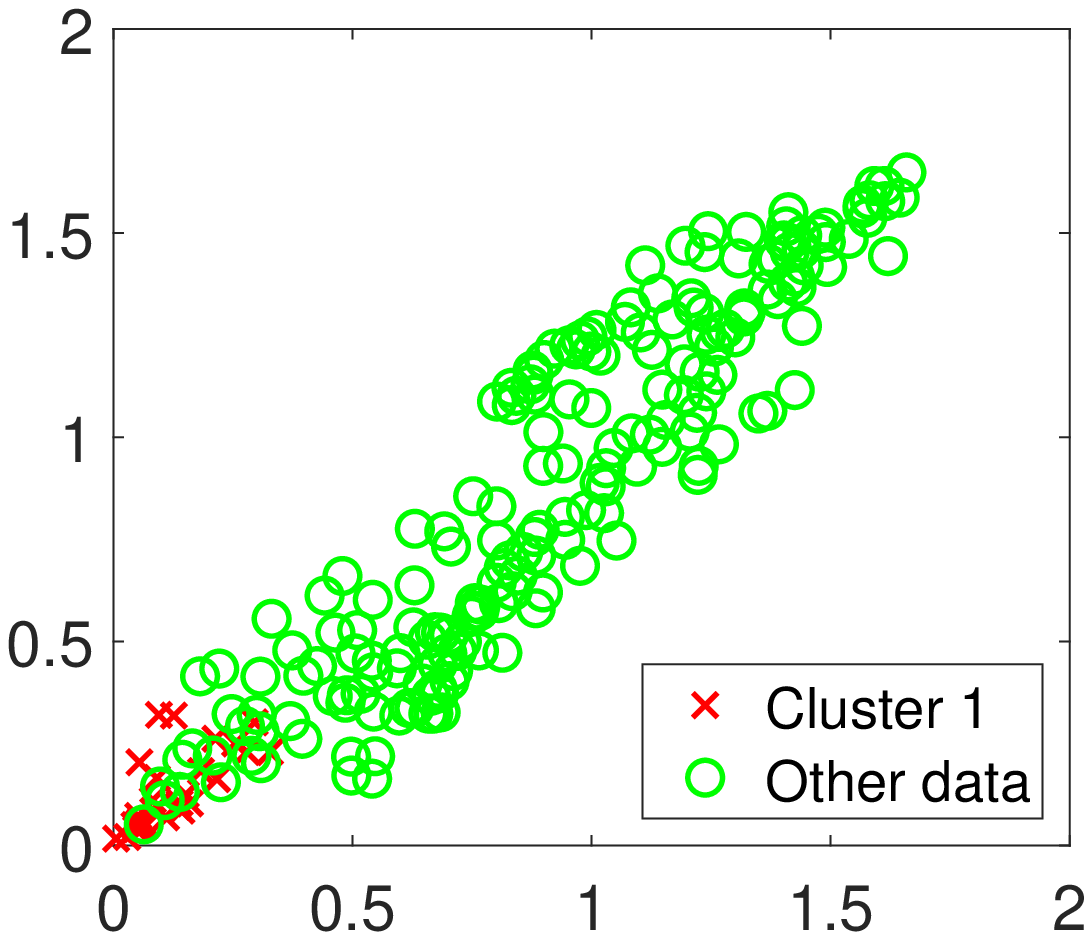}}
\subcaptionbox{\label{fig:watt1}}{
\includegraphics[width=0.48\linewidth]{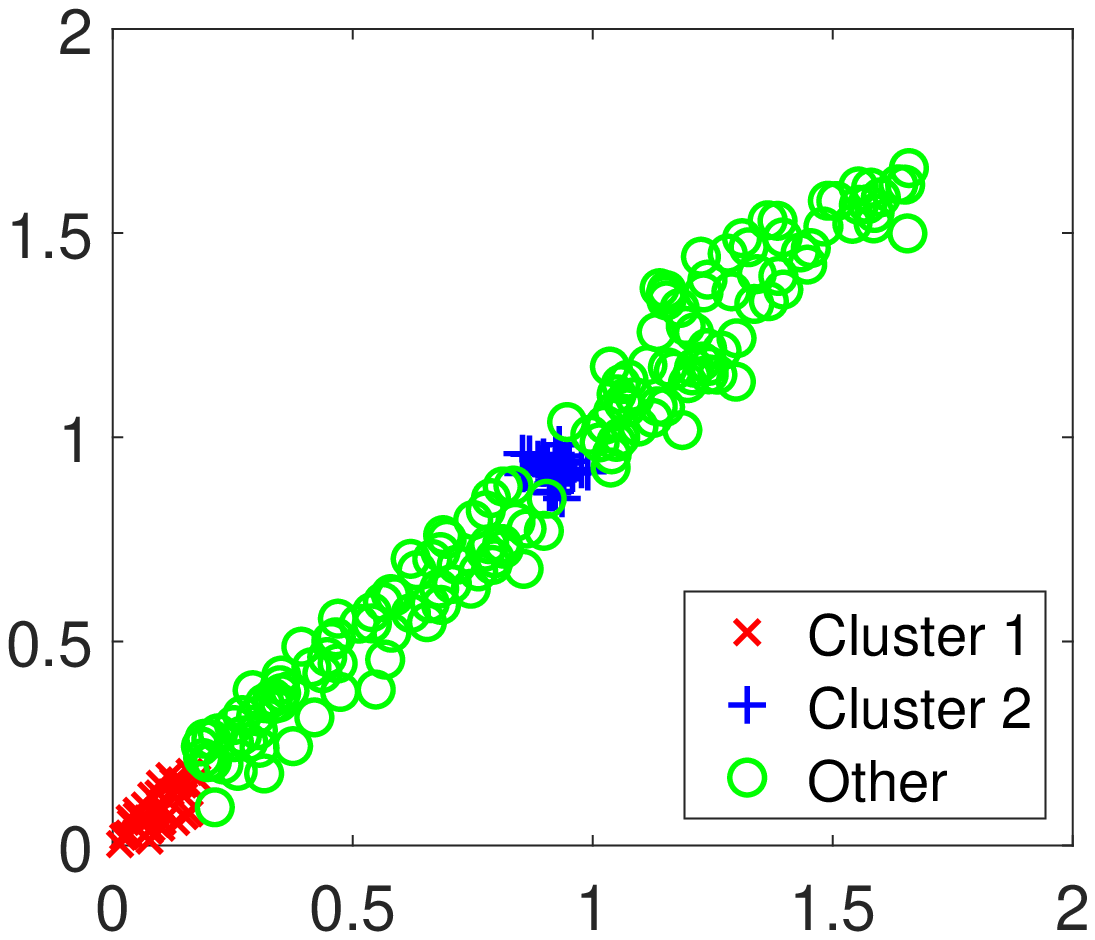}}
\subcaptionbox{\label{fig:woatt1}}{
\includegraphics[width=0.48\linewidth]{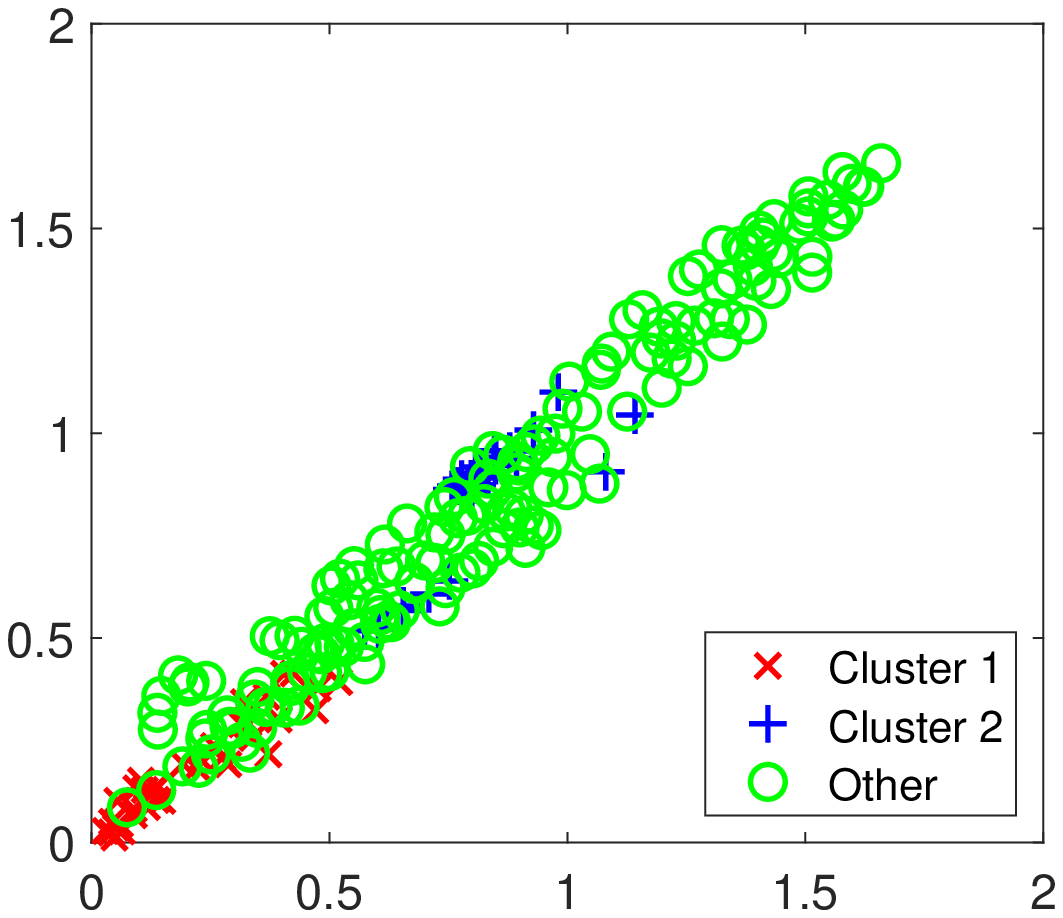}}
\caption{\small Visualization of 2D map of latent vector distribution in the static state. Case 1 with (a) cluster encoder generated by SCAE and (b) conventional encoder from autoencoder. Case 3 with (c) cluster encoder generated by SCAE and (d) conventional encoder from autoencoder. \label{fig:visualcompare2}}
\end{figure}

\subsubsection{Effect of Subcarrier Clustering}
The design of SCAE for subcarrier selection is an important component in our proposed ALPD system. The major design of SCAE for achieving better clustering performance is the additional inclusion of clustering loss $l^p_{clst}$ as defined in (\ref{lt}); whereas the conventional autoencoder only possesses the reconstruction loss of $l^p_{recst}$. To verify whether the designed clustering loss in SCAE can achieve desired performance, we set the length of latent vector $\boldsymbol{z}^p_{t,k,l}$ to be $2$ such that we can visually observe its data distribution by mapping the 2-dimensional vector elements into the horizontal and vertical axes.

Figs. \ref{fig:watt2} and \ref{fig:woatt2} demonstrate the respective comparison using the cluster encoder generated by SCAE and using the conventional encoder generated by autoencoder in Case 2 in Table \ref{case} with static state. Notice that the red clustered data points are classified by the encoder representing a special set of subcarriers that possess similar behaviors, i.e., a stationary person existing in Room 1, compared to those original green data points for the case in an empty room. It can be seen from Fig. \ref{fig:watt2} that all the red data points in Cluster 1 are aggregated at the lower-left corner by adopting the proposed SCAE scheme; whereas those are somewhat dispersed by adopting the conventional encoder as shown in Fig. \ref{fig:woatt2}. The effectiveness of SCAE scheme to provide a single effective aggregated cluster for Case 2 in static state can be therefore be revealed.

Furthermore, Figs. \ref{fig:watt1} and \ref{fig:woatt1} illustrate the comparison between SCAE and autoencoder with static state under Case 4, where there exist two stationary people with each of them in either one of the two rooms. It can be observed from both figures that there are two clusters generated named Cluster 1 and Cluster 2 respectively with red and blue data points in comparison with original green data points. The two distinct clusters imply two special groups of subcarriers with their own similar behaviors showing the signal characteristics of a stationary person existed in either Room 1 or Room 2. As can be observed from Fig. \ref{fig:watt1}, the two clusters generated by proposed SCAE possess two concentrated and distinct groups of data points; while the conventional autoencoder results in two confusing clusters overlapped by unrelated points as depicted in Fig. \ref{fig:woatt1}. Consequently, the SCAE scheme can feasibly construct two clusters with better clustering performance compared to conventional autoencoder.

\subsubsection{Relation Between Cluster and Weight}

\begin{figure}
\centering
\includegraphics[width=.5\textwidth]{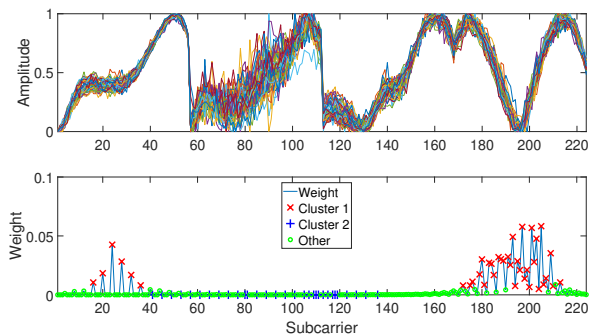}
\caption{\small Amplitude of the CSI subcarriers and the corresponding attention weights in different subcarrier clusters.}
\label{fig:weight1}
\end{figure}

\begin{figure}
\centering
\includegraphics[width=3.3in]{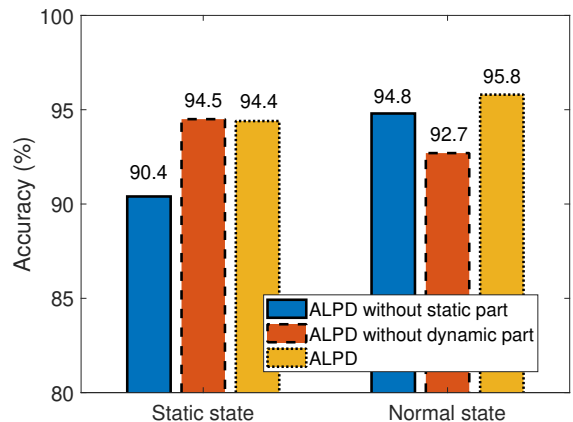}
\caption{\small Performance evaluation for ALPD in static and normal states.}
\label{fig:staticbar}
\end{figure}

\begin{figure*}[!ht]
\centering
\subcaptionbox{\label{fig:confusionpair1}}{\includegraphics[width=.32\linewidth]{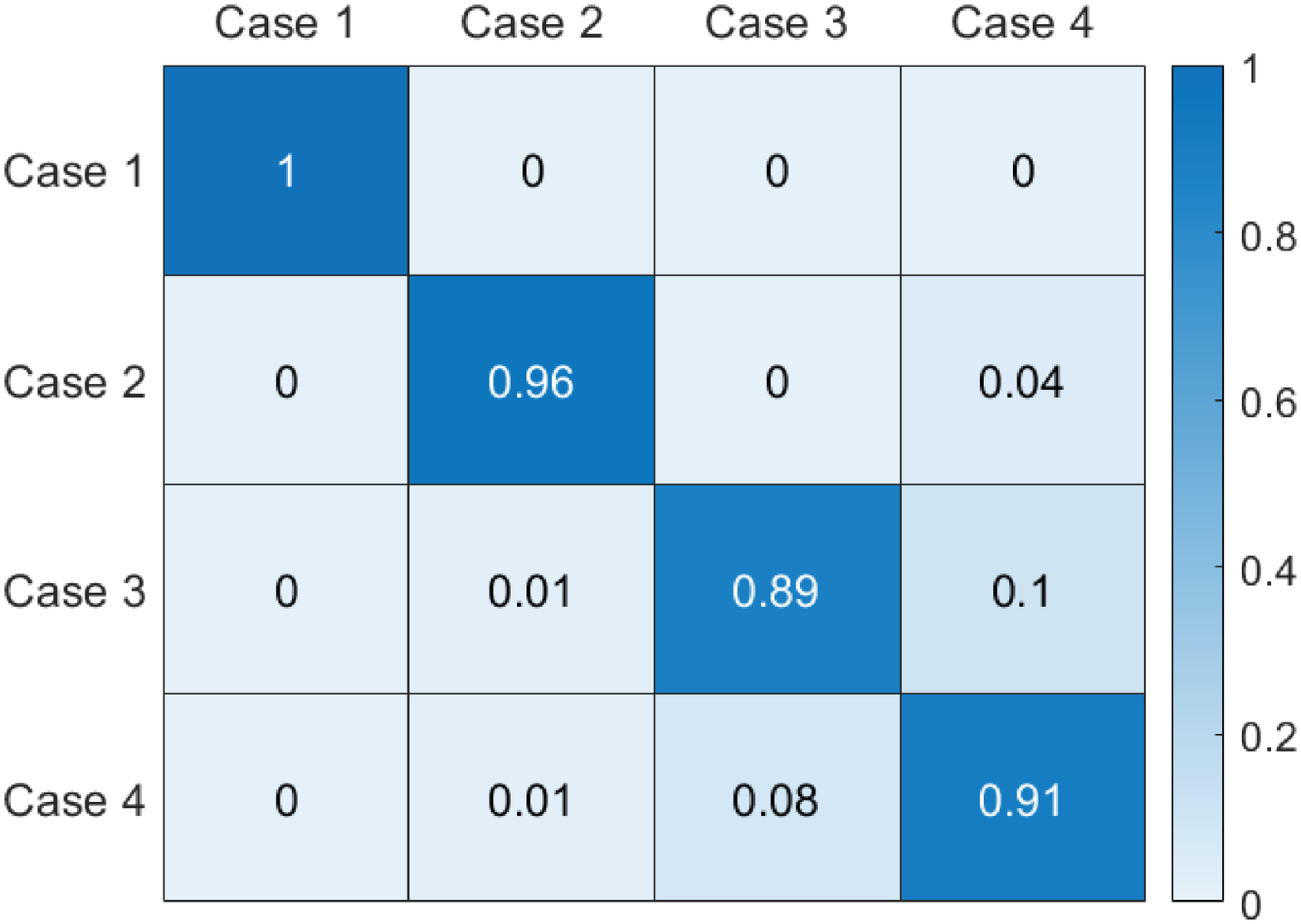}}
\subcaptionbox{\label{fig:confusionpair2}}{\includegraphics[width=.32\linewidth]{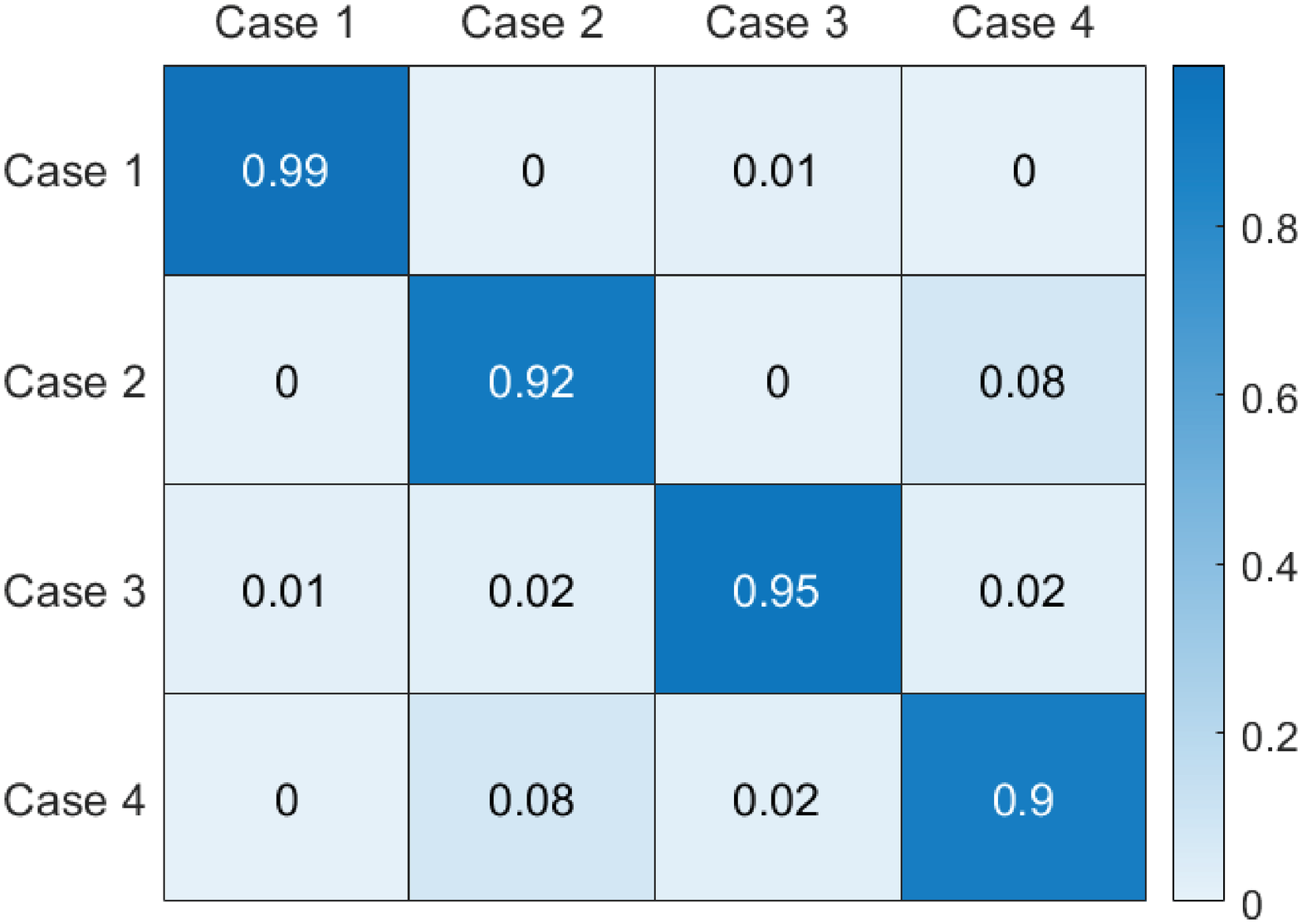}}
\subcaptionbox{\label{fig:confusionpair3}}{\includegraphics[width=.32\linewidth]{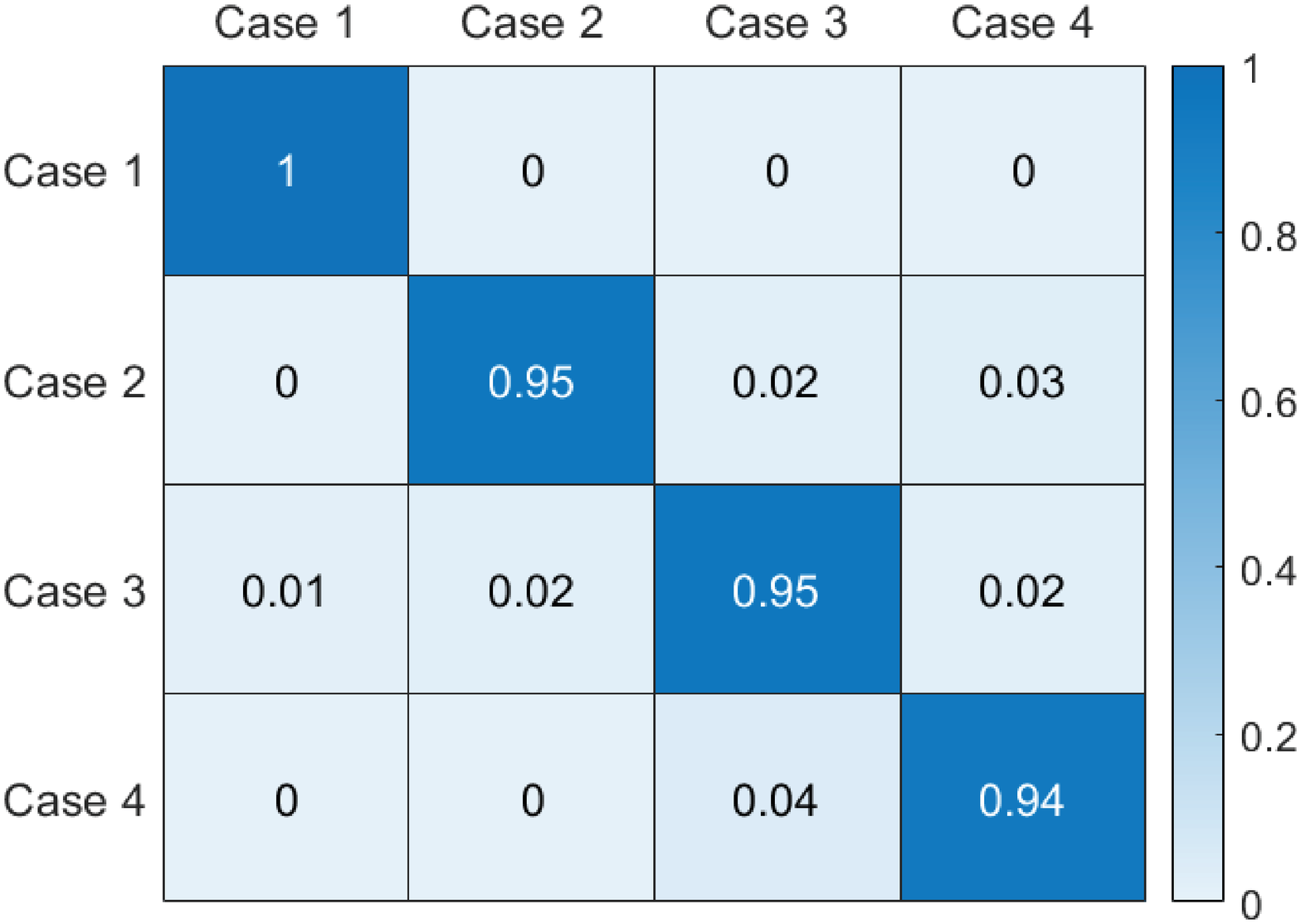}}
\caption{\small Confusion matrix of ALPD in the normal state using (a) pair 1, (b) pair 2 and (c) both pairs with average accuracy of $94\%$, $94\%$, and $96\%$, respectively.\label{fig:confusionpair}}
\end{figure*}

As mentioned in the previous subsection, the SCAE algorithm can cluster subcarriers with similar behavior from the normalized CSI amplitude feature vector of $\tilde{\boldsymbol{h}}^p_t$ in $\eqref{h_tilda}$ and $\boldsymbol{w}^p_{t,k,l}$ in $\eqref{ww}$. We can now investigate the usefulness of clustered data in attention-based subcarrier weighting, as shown in Fig. \ref{fig:subcarrierselection}. Fig. \ref{fig:weight1} displays the amplitude of CSI per subcarrier, as well as its corresponding extracted attention weight vector of $\boldsymbol{d}^p_t$ in $\eqref{dd}$. The results reveal that higher weights are assigned to subcarriers around $20$ to $40$ and from $170$ to $210$, indicating that informative features can be appropriately utilized for prediction. This observation also indicates that more stable CSI features are attainable in Cluster 1 compared to the trembling curves in Cluster 2, which are indexed from subcarriers $40$ to $140$. This outcome is reasonable because our attention-based subcarrier weighting aims to select relatively reliable subcarriers and assign them with higher weights.

\subsubsection{Performance of Subcarrier Selection}

Fig. \ref{fig:staticbar} depicts the performance of ALPD with/without static/dynamic feature parts in the static and normal states. When we employ ALPD without the static CSI feature, we can observe that it has the lowest accuracy of $90.4\%$ since the case of a standstill person in the room cannot be accurately detected. On the contrary, if we consider ALPD without dynamic CSI in state state, i.e., only with static feature, the highest accuracy is attained since the network is much overfitted to the dataset in static state. However, such overfitting issue takes place in the normal state containing the complete data, which leads to a performance degradation from $94.5\%$ to $92.7\%$. While, ALPD without static part is capable of estimating a more generic situation in normal state. Benefited by both static and dynamic designed in ALPD, it can achieve the highest accuracy of $95.8\%$ for dataset in normal state.

\subsection{Evaluation of Bidirectional Transmission}

\subsubsection{Complementarity of Bidirectional Transmission}

\begin{table}[]
\centering
\small
\caption{Relationship between Pair and Tx/Rx Deployment}
\begin{tabular}{|l|cc|cc|l|}
\hline
\multirow{2}{*}{Scenario} & \multicolumn{2}{l|}{Room 1}  & \multicolumn{2}{l|}{Room 2}  & \multirow{2}{*}{Note}                                                                         \\ \cline{2-5}
                          & \multicolumn{1}{l|}{Tx} & Rx & \multicolumn{1}{l|}{Tx} & Rx &                                                                                               \\ \hline
Pair 1                    & \multicolumn{1}{l|}{$\checkmark$}  &    & \multicolumn{1}{l|}{}   & $\checkmark$  & \begin{tabular}[c]{@{}l@{}}Presence in Tx for Case 2\\ Presence in Rx for Case 3\end{tabular} \\ \hline
Pair 2                    & \multicolumn{1}{l|}{}   & $\checkmark$  & \multicolumn{1}{l|}{$\checkmark$}  &    & \begin{tabular}[c]{@{}l@{}}Presence in Rx for Case 2\\ Presence in Tx for Case 3\end{tabular} \\ \hline
Both pairs                & \multicolumn{1}{l|}{$\checkmark$}  & $\checkmark$  & \multicolumn{1}{l|}{$\checkmark$}  & $\checkmark$  & Presence in both for all cases \\ \hline
\end{tabular}
\label{paircase}
\end{table}

As previously mentioned in Fig. \ref{observationFig}, the fluctuations of CSI caused by a person randomly walking in a room on the Rx side are more significant than those caused by a person on the Tx side. To study this phenomenon on detection accuracy, rather than using bidirectional transmission data, we consider to train ALPD with only one transmission pair. Table \ref{paircase} defines the relationship between the transmission pair and in four presence detection cases. Note that pair 1 indicates Tx and Rx APs are deployed in Rooms 1 and 2, respectively, and vice versa for transmission pair 2. Therefore, we can observe presence in Tx for Case 2 and Rx for Case 3 by using transmission transmission pair 1, and vice versa for dataset in transmission pair 2.

Figs. \ref{fig:confusionpair1} and \ref{fig:confusionpair2} illustrate the performance with a confusion matrix in the normal state corresponding to pairs 1 ($p=1$) and pair 2 ($p=2$), respectively. It can be observed that in Fig. \ref{fig:confusionpair1}, it becomes relatively difficult to distinguish human presence in Cases 3 and 4, resulting in a lower accuracy of $89\%$ and $91\%$. Similarly, in Fig. \ref{fig:confusionpair2}, Cases 2 and 4 have a comparably lower accuracy of $92\%$ and $90\%$. Table \ref{paircase} reveals that when a person is present at the Rx side, detecting human presence becomes a potential challenge. As previously explained, this phenomenon is due to the more severe fluctuations of CSI observed at the Rx side compared to the Tx side, making detection more difficult. However, with both pairs adopted in the proposed ALPD scheme, bidirectional transmission dataset can compromise such dilemma during training, as shown in Fig. \ref{fig:confusionpair3}. The asymmetry channel in presence detection is compensated by reciprocal CSI dataset. It demonstrates that ALPD with two pairs can achieve the highest average accuracy of $96\%$ compared to that employing either one of the two pairs.

\subsubsection{Effect of Interference Environment}

\begin{figure}
\centering
\includegraphics[width=3.3in]{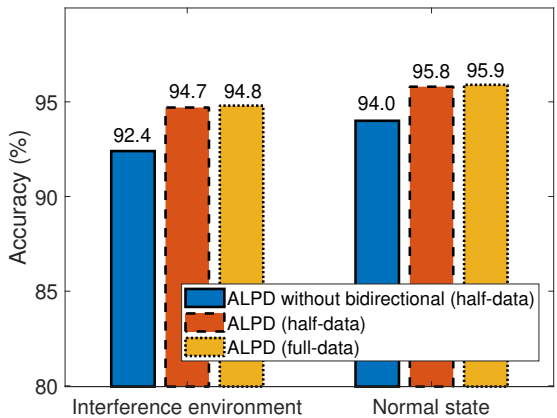}
\caption{\small Performance of ALPD in interference environment and normal states. Note that ALPD without bidirectional capability uses half of data for training, where the remaining ones employ either half or full dataset in both pairs.}
\label{fig:interferencebar}
\end{figure}

Fig. \ref{fig:interferencebar} illustrates the comparison between the accuracy of ALPD with and without bidirectional transmission data in the normal and interference states. We consider either full or half of training data to be applied for training APLD network. ALPD outperforms that without bidirectional transmission data with the accuracy difference of around $2.3\%$ and $1.8\%$ in interference and normal states, respectively. Benefited by bidirectional transmission data, it exhibits advantages of anti-interference and higher stability. To elaborate a little further, only $0.1\%$ accuracy difference is observed when employing half of data compared to full dataset, which implies that smaller amount of data can still effectively and efficiently detect the human presence in all cases.

\subsubsection{Amount of Utilized Training Data}

\begin{figure}
\centering
\includegraphics[width=3.3in]{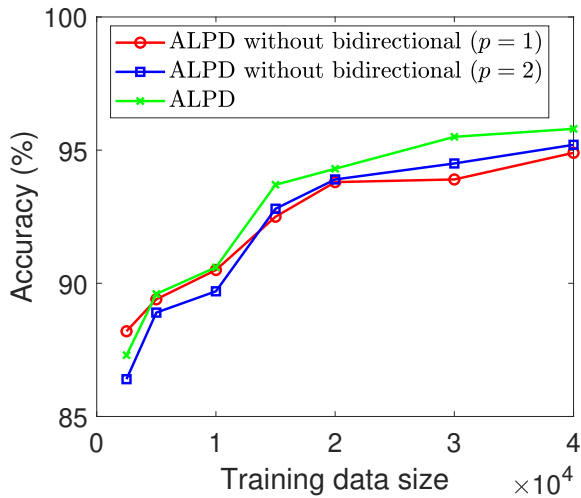}
\caption{\small Effect of different sizes of training data in ALPD with or without bidirectional capability.}
\label{fig:datausage}
\end{figure}

Fig. \ref{fig:datausage} illustrates the impact of the amount of training data on the accuracy of ALPD. It can be observed that when the full training data with an amount of 40000 is applied, ALPD outperforms the case without bidirectional dataset. Furthermore, the proposed ALPD only requires 25000 training data to achieve the same accuracy of $95\%$ as that without bidirectional dataset using 40000 training data. To elaborate further, pair 1 with less data outperforms pair 2, whereas pair 2 with more data achieves higher accuracy than pair 1. This implies that the case with Tx in Room 1 and Rx in Room 2 potentially benefits from low-cost training; however, overfitting issues may occur with more collected datasets. In a nutshell, the employment of bidirectional transmission data in ALPD is required to improve stability and accuracy as well as to reduce the costs of data collection for training.

\subsection{Benchmark Comparison}

\begin{figure}
\centering
\includegraphics[width=3.5in]{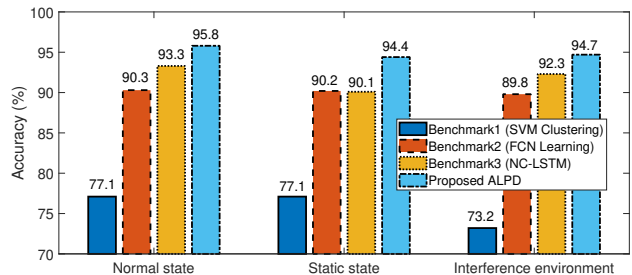}
\caption{\small Performance of the proposed ALPD scheme compared to the existing methods in cases of normal, static and interference states.}
\label{fig:totalbar}
\end{figure}

Fig. \ref{fig:totalbar} presents a comparison of our ALPD system with existing benchmarks, including SVM-based CSI detection \cite{pd_svm}, a multi-layer FCN-based structure \cite{dnn_pd}, and a non-collaborative (NC) LSTM method \cite{f_lstm}, in normal, static, and interference states. To ensure a fair comparison, we train the benchmarks using SVM clustering, FCN, and NC-LSTM using all the training data of a single transmission pair $p=1$, based on their deployment assumptions. In contrast, ALPD is trained with two transmission pairs using the same amount of total data. As depicted in Fig. \ref{fig:totalbar}, the accuracy of ALPD outperforms benchmarks 1 to 3 by $18.7\%$, $5.5\%$, and $2.5\%$, respectively in normal state. This is because SVM with non-deep neural networks is unable to perform appropriate clustering to those non-linear, non-convex, as well as irregular CSI dataset, especially for the data with human presence. Moreover, benchmark 2 using FCN can only extract potential huge difference between empty and rooms with people presence, leading to a comparably lower accuracy that NC-LSTM. On the other hand, NC-LSTM focuses on the temporal feature rather than the spectral and spatial information and has a lower accuracy than our proposed ALPD system by more than $2\%$ in all three states. In the static state, where fewer subcarriers are affected by small fluctuations in CSI, ALPD achieves $94.4\%$ accuracy with the attention mechanism generating significant weights for subcarriers to be selected. In contrast, benchmarks 2 and 3 have asymptotic performances in the static state. By comparing the normal and interference states, we observe that the accuracy declines by around $1\%$ to $4\%$ when people are in the interference zone. This is because the CSI signals may pass through the walls to the other space, which cannot be estimated. Without the compensation of bi-directional transmission pairs, benchmarks 1 to 3 cannot properly conduct human presence detection in interference state. The proposed ALPD with bi-directional transmission APs achieves the highest accuracy of $94.7\%$ in interference environments with the aid of extracted spectral, spatial, and temporal features.

\begin{table}[]
\centering
\scriptsize
\caption{Human Activity Detection Cases}
\begin{tabular}{|l|cccc|cccc|}
\hline
 & \multicolumn{4}{l|}{Room 1} & \multicolumn{4}{l|}{Room 2} \\ \hline
 & \multicolumn{1}{l|}{Empty} & \multicolumn{1}{l|}{Walk} & \multicolumn{1}{l|}{Run} & Jump & \multicolumn{1}{l|}{Empty} & \multicolumn{1}{l|}{Walk} & \multicolumn{1}{l|}{Run} & Jump \\ \hline
Case 1 & \multicolumn{1}{l|}{$\checkmark$} & \multicolumn{1}{l|}{} & \multicolumn{1}{l|}{} &  & \multicolumn{1}{l|}{$\checkmark$} & \multicolumn{1}{l|}{} & \multicolumn{1}{l|}{} &  \\ \hline
Case 2 & \multicolumn{1}{l|}{} & \multicolumn{1}{l|}{$\checkmark$} & \multicolumn{1}{l|}{} &  & \multicolumn{1}{l|}{$\checkmark$} & \multicolumn{1}{l|}{} & \multicolumn{1}{l|}{} &  \\ \hline
Case 3 & \multicolumn{1}{l|}{} & \multicolumn{1}{l|}{} & \multicolumn{1}{l|}{$\checkmark$} &  & \multicolumn{1}{l|}{$\checkmark$} & \multicolumn{1}{l|}{} & \multicolumn{1}{l|}{} &  \\ \hline
Case 4 & \multicolumn{1}{l|}{} & \multicolumn{1}{l|}{} & \multicolumn{1}{l|}{} & $\checkmark$ & \multicolumn{1}{l|}{$\checkmark$} & \multicolumn{1}{l|}{} & \multicolumn{1}{l|}{} &  \\ \hline
Case 5 & \multicolumn{1}{l|}{$\checkmark$} & \multicolumn{1}{l|}{} & \multicolumn{1}{l|}{} &  & \multicolumn{1}{l|}{} & \multicolumn{1}{l|}{$\checkmark$} & \multicolumn{1}{l|}{} &  \\ \hline
Case 6 & \multicolumn{1}{l|}{$\checkmark$} & \multicolumn{1}{l|}{} & \multicolumn{1}{l|}{} &  & \multicolumn{1}{l|}{} & \multicolumn{1}{l|}{} & \multicolumn{1}{l|}{$\checkmark$} &  \\ \hline
Case 7 & \multicolumn{1}{l|}{$\checkmark$} & \multicolumn{1}{l|}{} & \multicolumn{1}{l|}{} &  & \multicolumn{1}{l|}{} & \multicolumn{1}{l|}{} & \multicolumn{1}{l|}{} & $\checkmark$ \\ \hline
Case 8 & \multicolumn{1}{l|}{} & \multicolumn{1}{l|}{$\checkmark$} & \multicolumn{1}{l|}{} &  & \multicolumn{1}{l|}{} & \multicolumn{1}{l|}{$\checkmark$} & \multicolumn{1}{l|}{} &  \\ \hline
Case 9 & \multicolumn{1}{l|}{} & \multicolumn{1}{l|}{} & \multicolumn{1}{l|}{$\checkmark$} &  & \multicolumn{1}{l|}{} & \multicolumn{1}{l|}{} & \multicolumn{1}{l|}{$\checkmark$} &  \\ \hline
Case 10 & \multicolumn{1}{l|}{} & \multicolumn{1}{l|}{} & \multicolumn{1}{l|}{} & $\checkmark$ & \multicolumn{1}{l|}{} & \multicolumn{1}{l|}{} & \multicolumn{1}{l|}{} & $\checkmark$ \\ \hline
\end{tabular} \label{casesact}
\end{table}

\begin{figure*}[!t]
\centering
\subcaptionbox{\label{fig:act1}}{\includegraphics[width=.325\linewidth]{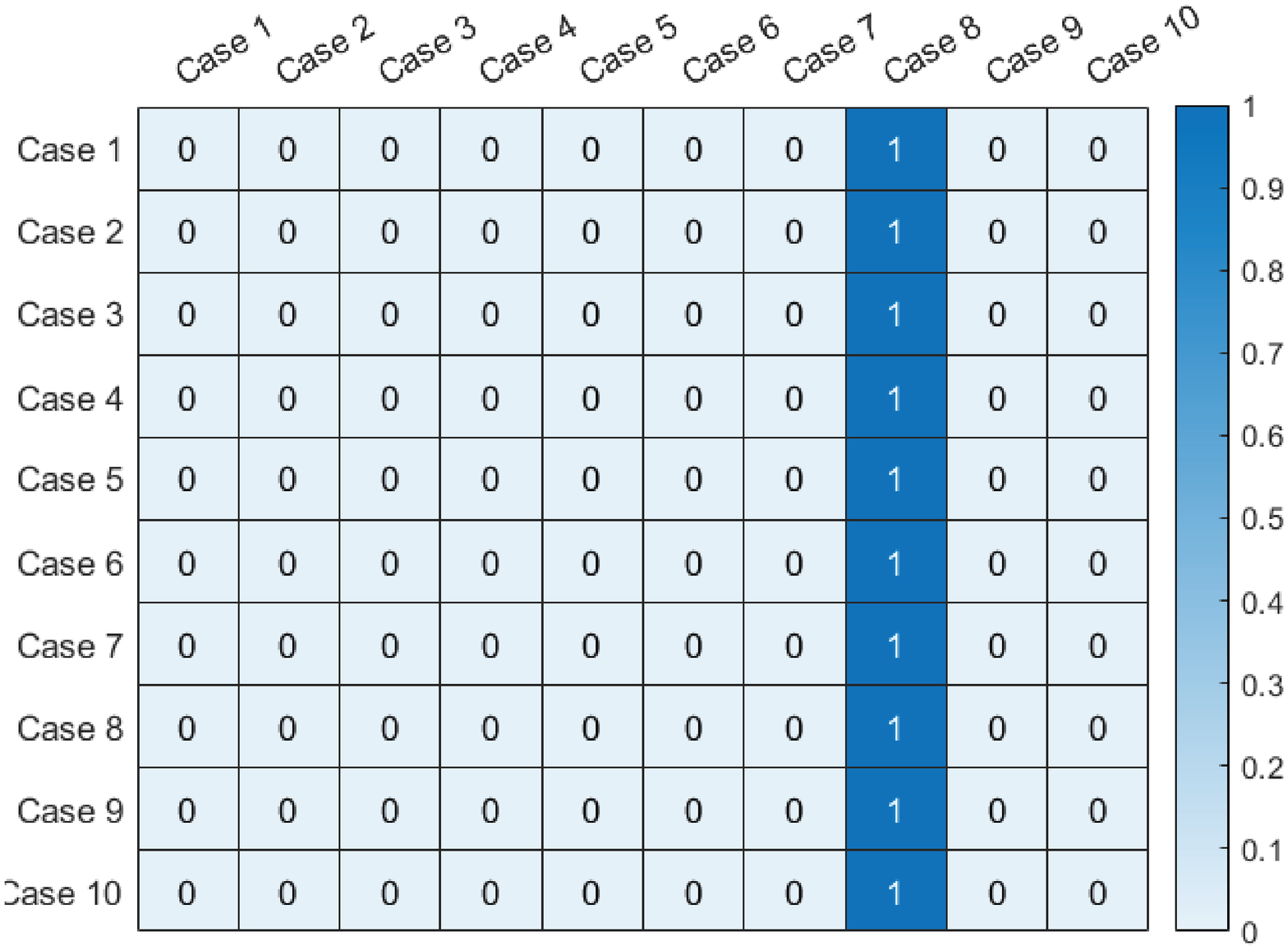}}
\subcaptionbox{\label{fig:act2}}{\includegraphics[width=.325\linewidth]{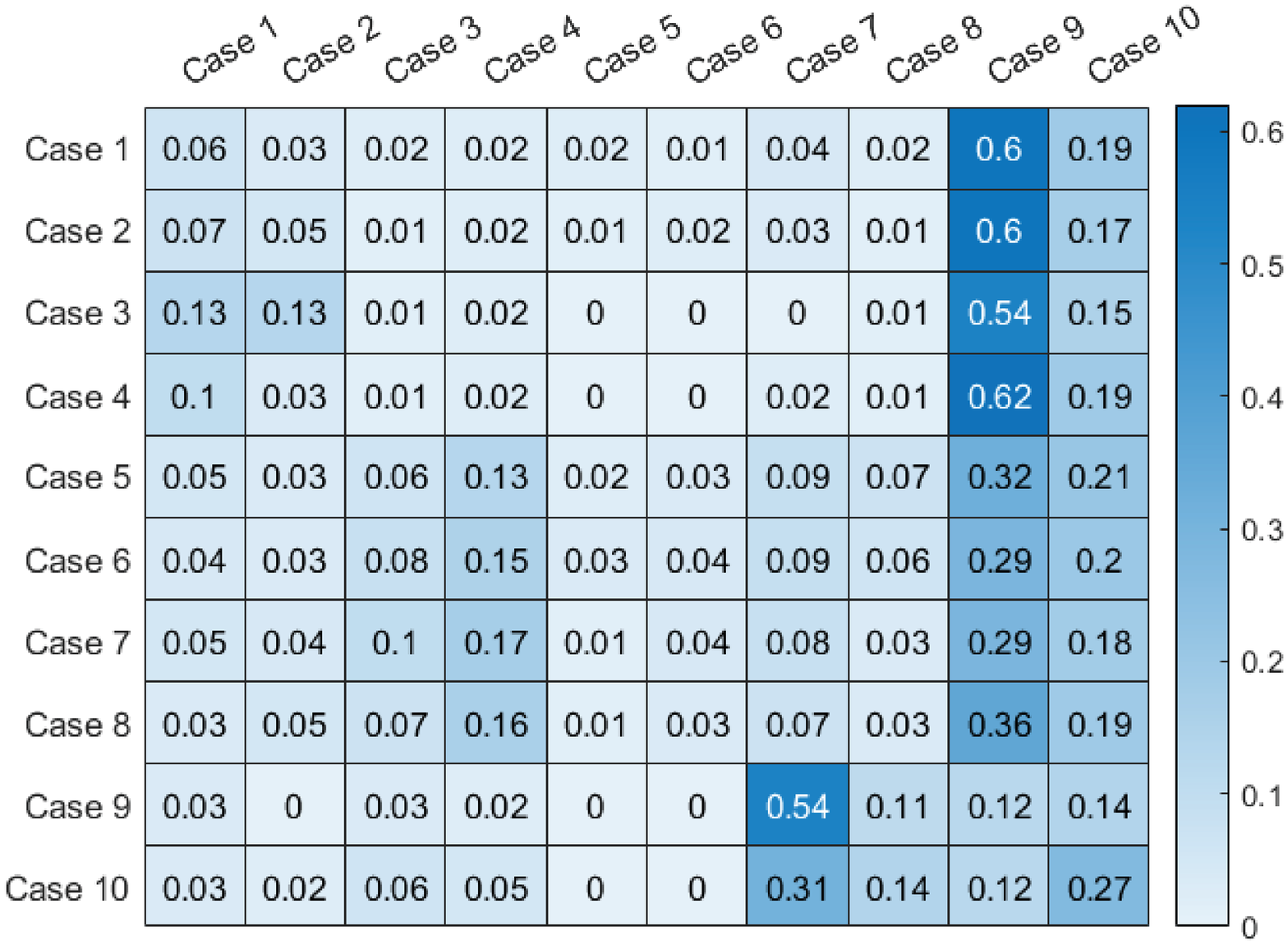}}
\subcaptionbox{\label{fig:act3}}{\includegraphics[width=.325\linewidth]{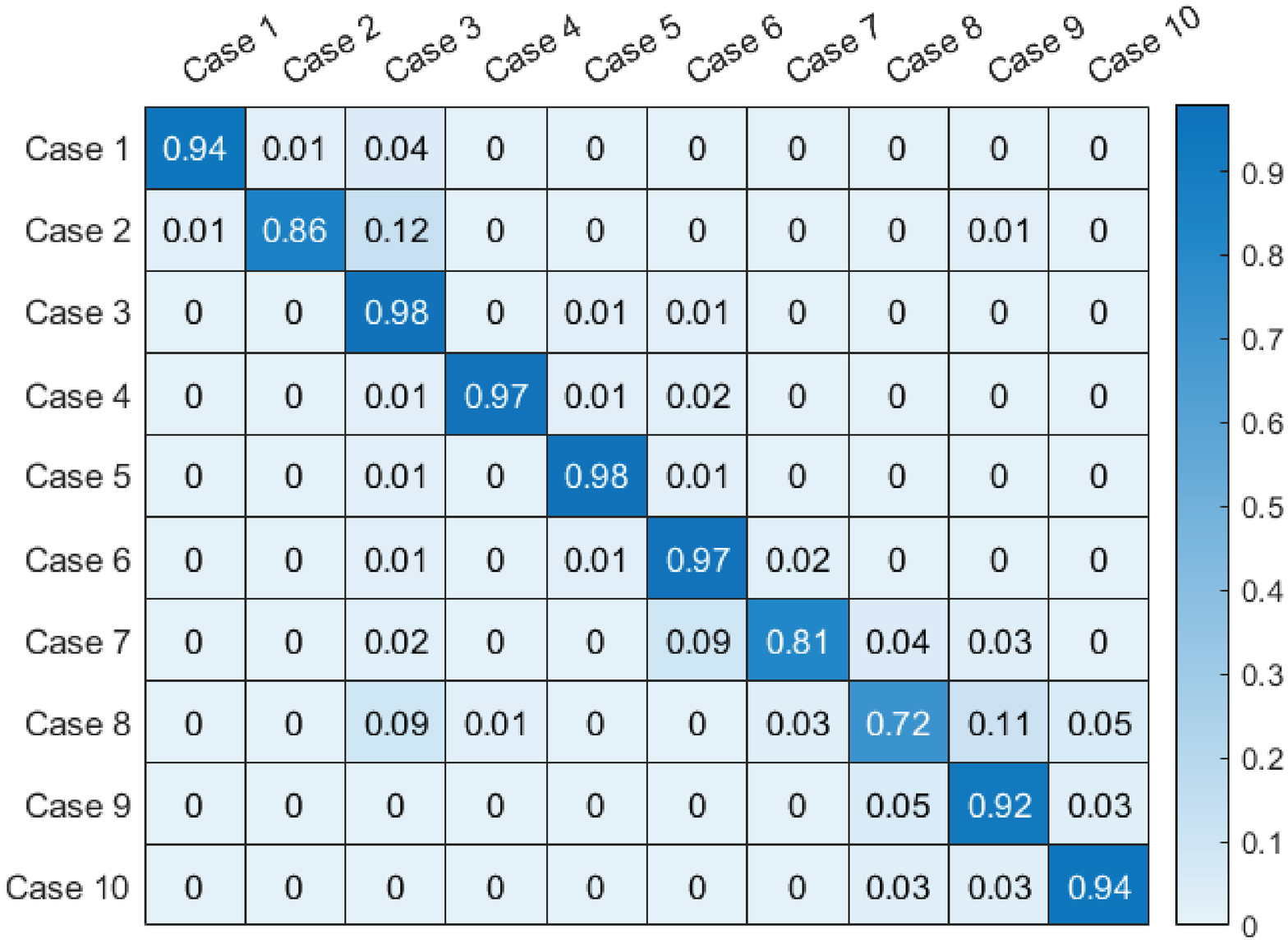}}
\caption{\small Confusion matrix of ten human activity detection cases with (a) benchmark 2, (b) benchmark 3, and (c) proposed ALPD scheme, having average accuracy of $10\%$, $7\%$, and $90.9\%$, respectively.\label{fig:act}}
\end{figure*}

Moreover, we have also evaluated the potentials of the proposed ALPD scheme in human activity detection in multi-rooms. As listed in Table \ref{casesact}, we consider ten cases with combinations of empty rooms and human walking, running as well as jumping. Note that Case 1 represents both rooms being empty. On the other hand, Cases 2 to 7 indicate activity in one room, while the other room remains empty. Cases 8 to 10 account for both rooms having the same type of activity. In Fig. \ref{fig:act}, we have compared the result of the proposed ALPD scheme to benchmarks 2 and 3. Owing to simple spatial and temporal feature extraction mechanisms adopted, we can observe that both benchmarks have compellingly lower accuracy compared to the proposed ALPD scheme, i.e., they are untrainable and predict by guessing the cases. In other words, benchmark 2 with spatial feature extraction cannot extract much smaller difference in activity detection compared to the cases between empty rooms and human presence. On the other hand, benchmark 3 only extracts temporal features rather than spatial features. Moreover, since both benchmarks 2 and 3 did not perform subcarrier selection, some less-informative subcarriers may dominate the hidden features, deteriorating the prediction result. However, ALPD is capable of achieving an human activity detection accuracy around $90.9\%$ thanks to joint benefits from the benchmarks and latent clustering, bidirectional feature extraction as well as attention-base subcarrier selection. Additionally, we can observe that the average accuracy of ALPD degrades from $95.8\%$ in Fig. \ref{fig:totalbar} to $90.9\%$ in Fig. \ref{fig:act3} due to more fine-grained cases to be detected. Furthermore, it can inferred from Fig. \ref{fig:act3} that cases of both rooms with activities have lower accuracy than either one room being empty, since it is more challenging to be detect two people with highly variant CSI. To elaborate a little further, it requires laborious data collection with the exponentially increasing numbers of activities and rooms. Therefore, efficient detection and data collection are regarded as essential designs in the future.

\section{Conclusion} \label{con}

In this paper, we have conceived an ALPD system using WiFi CSI signals that leverages a combination of spectral, spatial, and temporal features with attention-based subcarrier weighting mechanism. ALPD achieves state-of-the-art accuracy in human presence detection in various states, including normal, static, and interference environments. The employment of bidirectional transmission data in ALPD is required to improve stability and accuracy, as well as to reduce the costs of data collection for training. Additionally, we demonstrate that using clustered data in attention-based subcarrier weighting improves the accuracy of human presence detection. To elaborate a little further, we have evaluated the potential of ALPD for detecting fine-grained human activities in multi-rooms. In conclusion, our proposed ALPD system outperforms the other existing benchmarks in open literature, which are designed based on SVM, FCN and LSTM methods.

%\end{document}

%\linespread{0.95}
\footnotesize
\bibliographystyle{IEEEtran}
\bibliography{myReference}

\end{document}